\documentclass[twocolumn]{IEEEtran}

\usepackage[usenames,dvipsnames,svgnames,table]{xcolor}
\usepackage{wrapfig}
\usepackage{times}
\usepackage{multicol}
\usepackage{multirow}
\usepackage[bookmarks=true, draft]{hyperref}
\usepackage{graphicx}
\usepackage{amsmath,amsthm,amssymb}
\usepackage{pgf,tikz}
\usepackage{mathrsfs}
\usepackage[linesnumbered,ruled,vlined]{algorithm2e}
\usepackage{epstopdf}
\usepackage{overpic}
\usepackage{xcolor}
\usepackage{xargs} 
\usepackage{verbatim} 
\usepackage[textsize=footnotesize]{todonotes}
\usepackage{enumerate}
\usepackage{placeins} 
\usepackage{marvosym}
\usepackage{tabularx}

\def\citep{\cite}

\tikzset{every node/.style={thick,shape=circle,draw=black,fill=blue!15,text=black,minimum size=1.7em,inner sep=0.5}}
\tikzset{every picture/.style=thick}
\colorlet{customLightRed}{red!65}     

\newcommand{\Wspace}{{\mathcal{W}}}
\newcommand{\objects}{{\mathcal{O}}}
\newcommand{\actions}{{\mathcal{A}}}
\newcommand{\Cspace}{{\mathcal{C}}}
\newcommand{\Fspace}{{\mathcal{F}}}

\newcommand{\toro}{{\tt TORO}\xspace}

\newcommand{\rno}{{\tt TORO-NO}\xspace}
\newcommand{\uno}{{\tt TORO-UNO}\xspace}
\newcommand{\rwo}{{\tt TORO}\xspace}
\newcommand{\tsp}{{\tt TSP}\xspace}

\newcommand{\fvs}{{\tt FVS}\xspace}
\def\rnos{{\sc ToroNoTSP}\xspace}
\def\lwoalgs{{\sc ToroFVSSingle}\xspace}

\newcommand{\DAG}{DAG\xspace}

\makeatletter
\def\thm@space@setup{\thm@preskip=0pt
\thm@postskip=0pt}
\makeatother

\newcommandx{\nt}[2][1=]{\todo[linecolor=blue,
			backgroundcolor=blue!10,bordercolor=blue,#1]{#2}}

\newtheorem{problem}{Problem}
\newtheorem{lemma}{Lemma}[section]
\newtheorem{observation}{Observation}[section]

\newtheorem{theorem}{Theorem}[section]
\theoremstyle{definition}

\newtheorem{remark}{Remark}

{\end{list}}

\newcolumntype{Z}{>{\centering\let\newline\\\arraybackslash\hspace{0pt}}X}

\renewcommand{\copyright}{\textsuperscript{\textregistered}}

\begin{document}


\title{Complexity Results and Fast Methods for Optimal Tabletop Rearrangement with Overhand Grasps}

\author{Shuai D Han,
        Nicholas M Stiffler,
        Athanasios Krontiris, 
        Kostas E Bekris, and
        Jingjin Yu%
}

%
%
\maketitle

%
%
\begin{abstract}
This paper studies the underlying combinatorial structure of a class
of object rearrangement problems, which appear frequently in
applications. The problems involve multiple, similar-geometry objects
placed on a flat, horizontal surface, where a robot can approach them
from above and perform pick-and-place operations to rearrange them.
The paper considers both the case where the start and goal object
poses overlap, and where they do not.  For overlapping poses, the
primary objective is to minimize the number of pick-and-place actions
and then to minimize the distance traveled by the end-effector. For
the non-overlapping case, the objective is solely to minimize the
travel distance of the end-effector. While such problems do not
involve all the complexities of general rearrangement, they remain
computationally hard in both cases.  This is shown through reductions
from well-understood, hard combinatorial challenges to these
rearrangement problems. The reductions are also shown to hold in the
reverse direction, which enables the convenient application on
rearrangement of well studied algorithms. These algorithms can be
very efficient in practice despite the hardness results. The paper
builds on these reduction results to propose an algorithmic pipeline
for dealing with the rearrangement problems.  Experimental evaluation,
including hardware-based trials, shows that the proposed pipeline
computes high-quality paths with regards to the optimization
objectives.  Furthermore, it exhibits highly desirable scalability as
the number of objects increases in both the overlapping and
non-overlapping setup.
\end{abstract}

\section{Introduction}
\label{sec:introduction}

\begin{figure}[t]
  \centering
   \includegraphics[width = 0.49 \columnwidth, height = 1.2in]{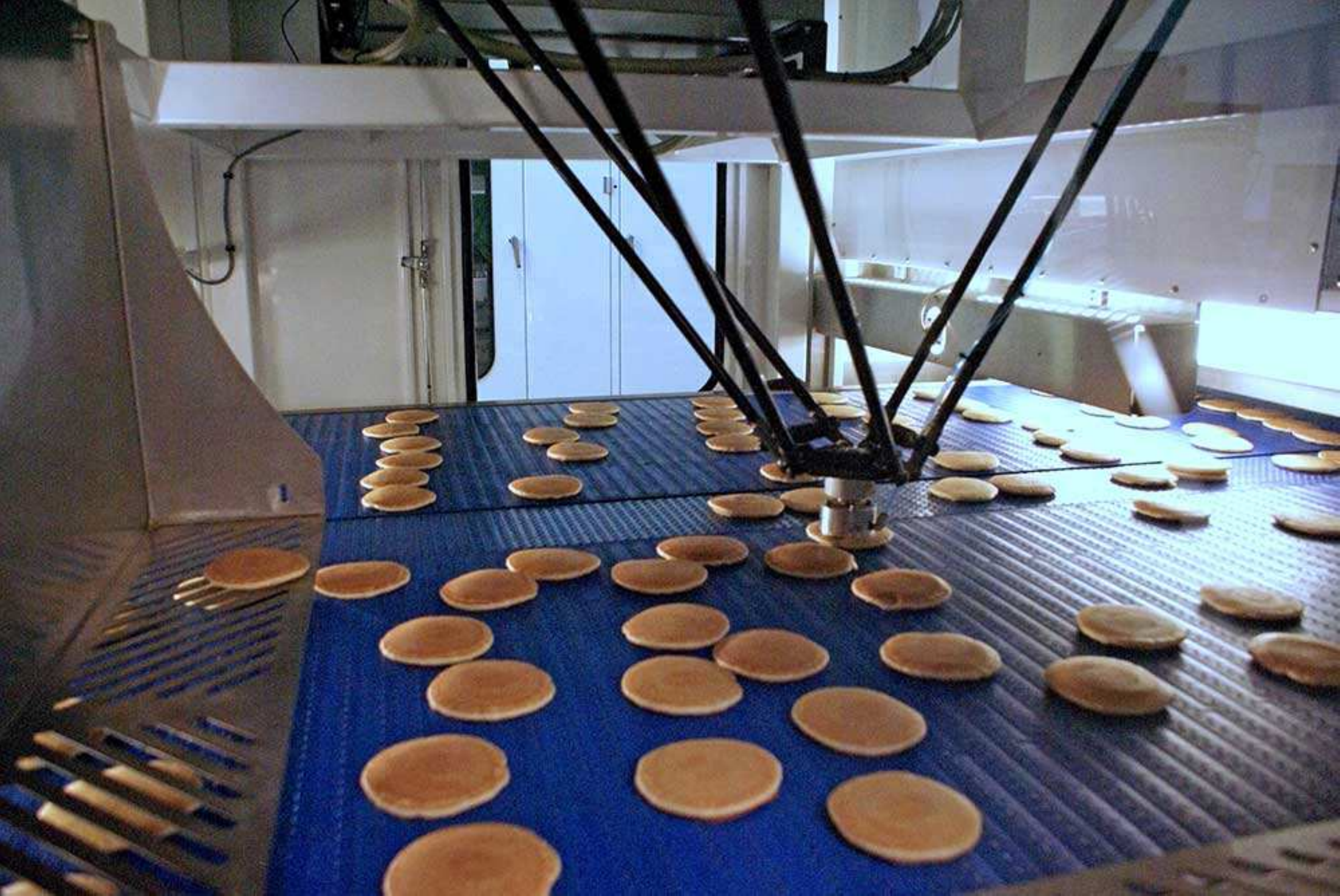}
   \includegraphics[width = 0.49 \columnwidth, height = 1.2in]{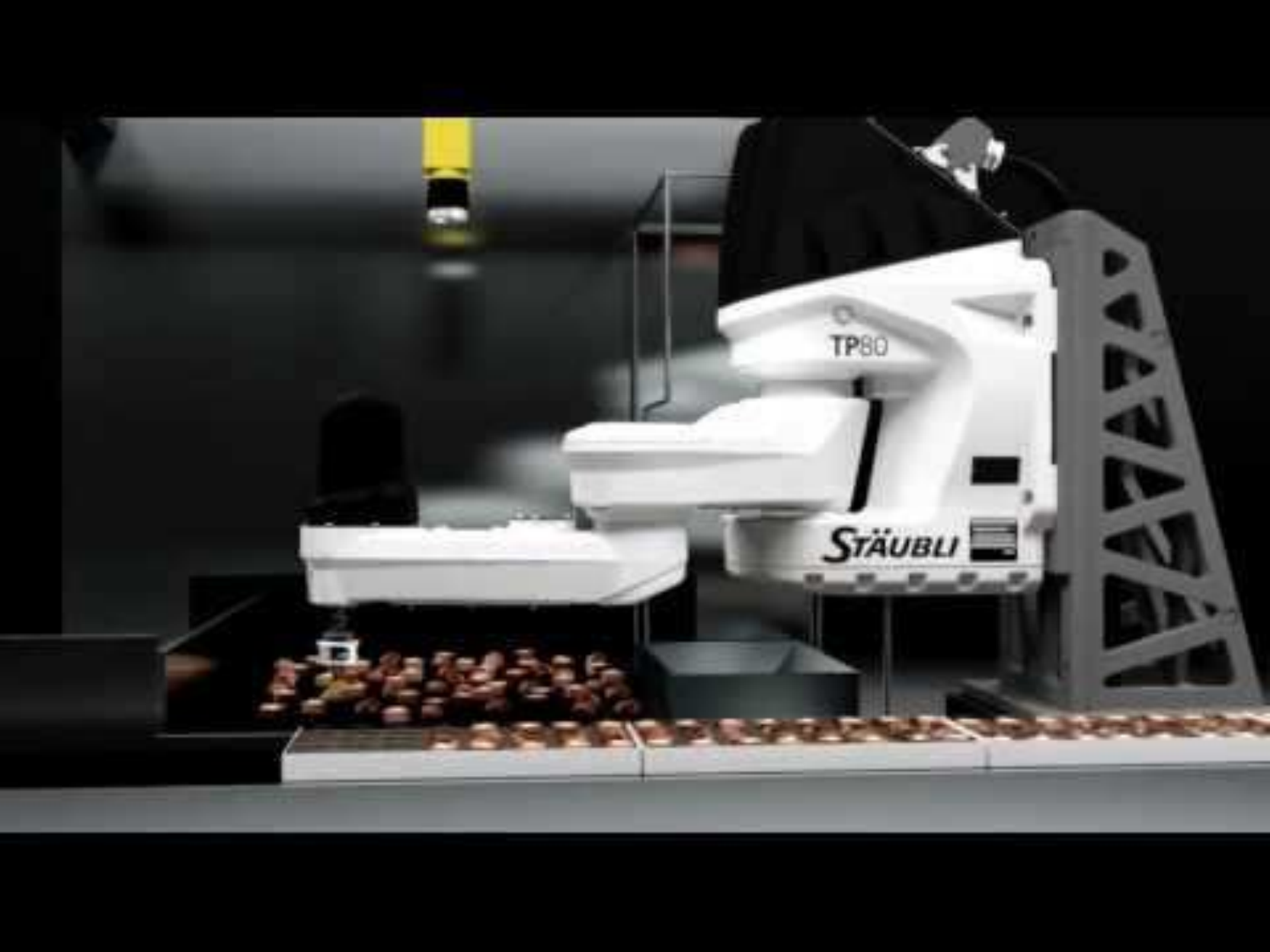}
 \caption{Examples of robots deployed in industrial settings tasked to
   arrange objects in desired configurations through pick and place:
   (left) ABB's IRB 360 FlexPicker rearranging pancakes (right)
   St\"{a}ubli's TP80 Fast Picker robot.}
 \label{fig:automation}
\end{figure}

In many industrial and logistics applications, such as those shown in
Fig. \ref{fig:automation}, a robot is tasked to rearrange multiple,
similar objects placed on a tabletop into a desired arrangement. In
these setups, the robot needs to approach the objects from above and
perform a \textit{pick-and-place action} at desired target poses. Such
operations are frequently part of product packaging and inspection
processes. Efficiency plays a critical role in these domains, as the
speed of task completion has a direct impact on financial viability;
even a marginal improvement in productivity could provide significant
competitive advantage in practice.  Beyond industrial robotics, a home
assistant robot may need to deal with such problems as part of a room
cleaning task. The reception of such a robot by people will be more
positive if its solutions are efficient and the robot does not waste
time performing redundant actions. Many subtasks affect the efficiency
of the overall solution in all of these applications, ranging from
perception to the robot's speed in grasping and transferring
objects. But overall efficiency also critically depends on the
underlying combinatorial aspects of the problem, which relate to the
number of pick-and-place actions that the robot performs, the
placement of the objects, as well as the sequence of objects
transferred.

This paper deals with the combinatorial aspects of tabletop object
rearrangement tasks. The objective is to understand the underlying
structure and obtain high-quality solutions in a computationally
efficient manner. The focus is on a subset of general rearrangement
problems, which relate to the above mentioned applications. In
particular, the setup corresponds to rearranging multiple,
similar-geometry, non-stacked objects on a flat, horizontal surface
from given initial to target arrangements. The robot can approach the
objects from above, pick them up and raise them. At that point, it can
move them freely without collisions with other objects.

\begin{figure*}[t]
    \centering
    \begin{tabular}{@{}cccc@{}}
        \includegraphics[height=1.3in]{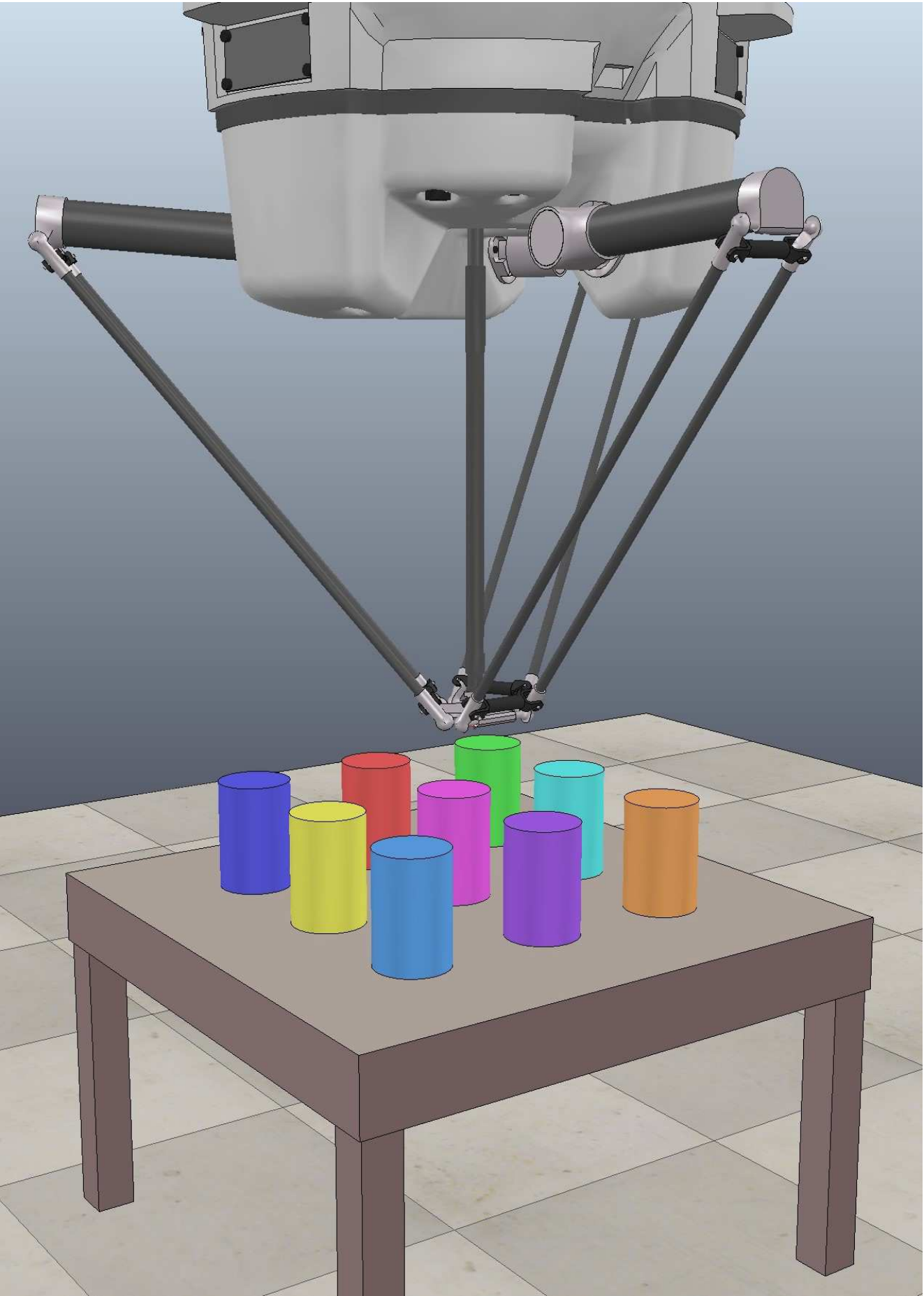} &
        \includegraphics[height=1.3in]{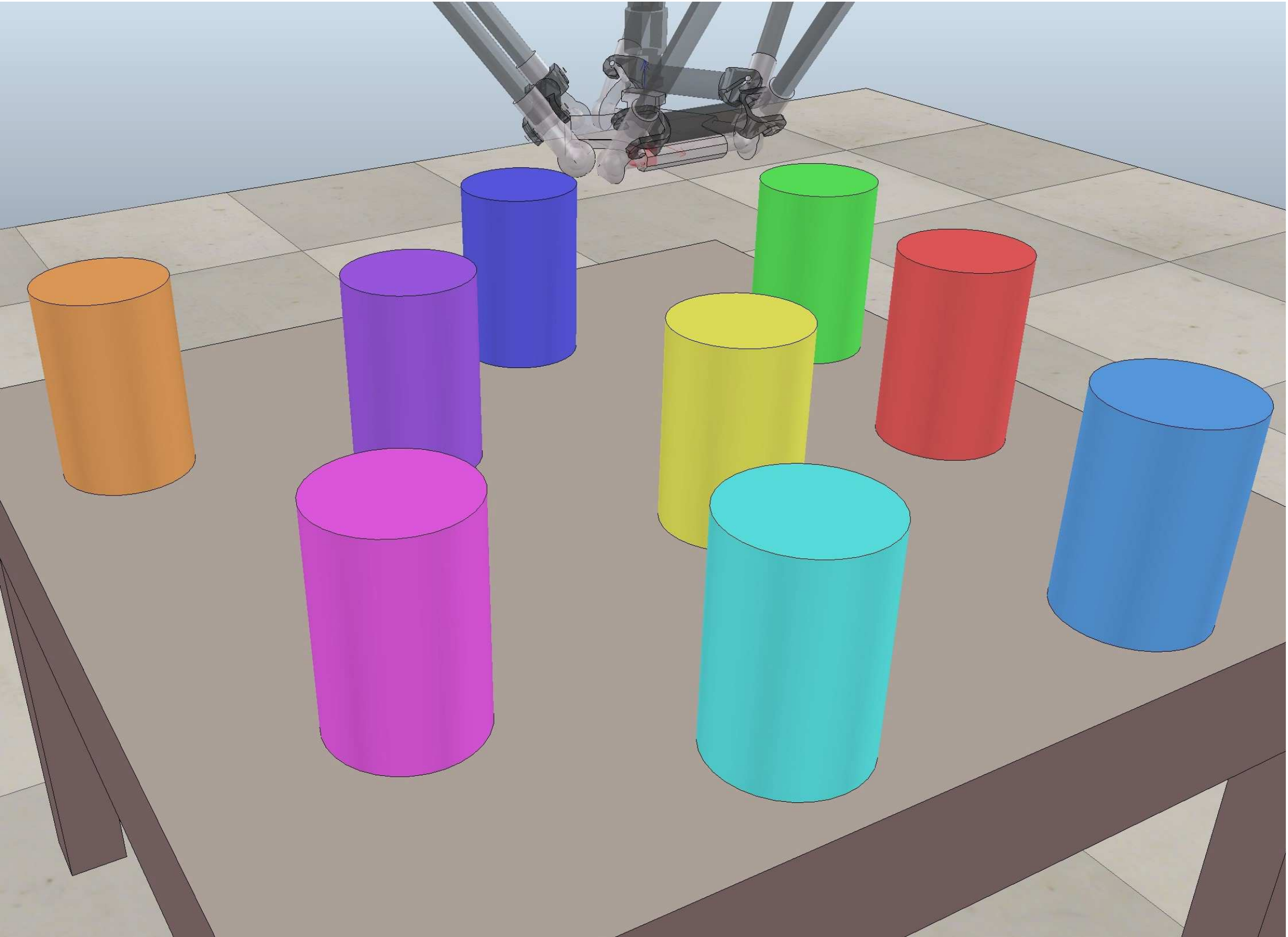} &
        \includegraphics[height=1.3in]{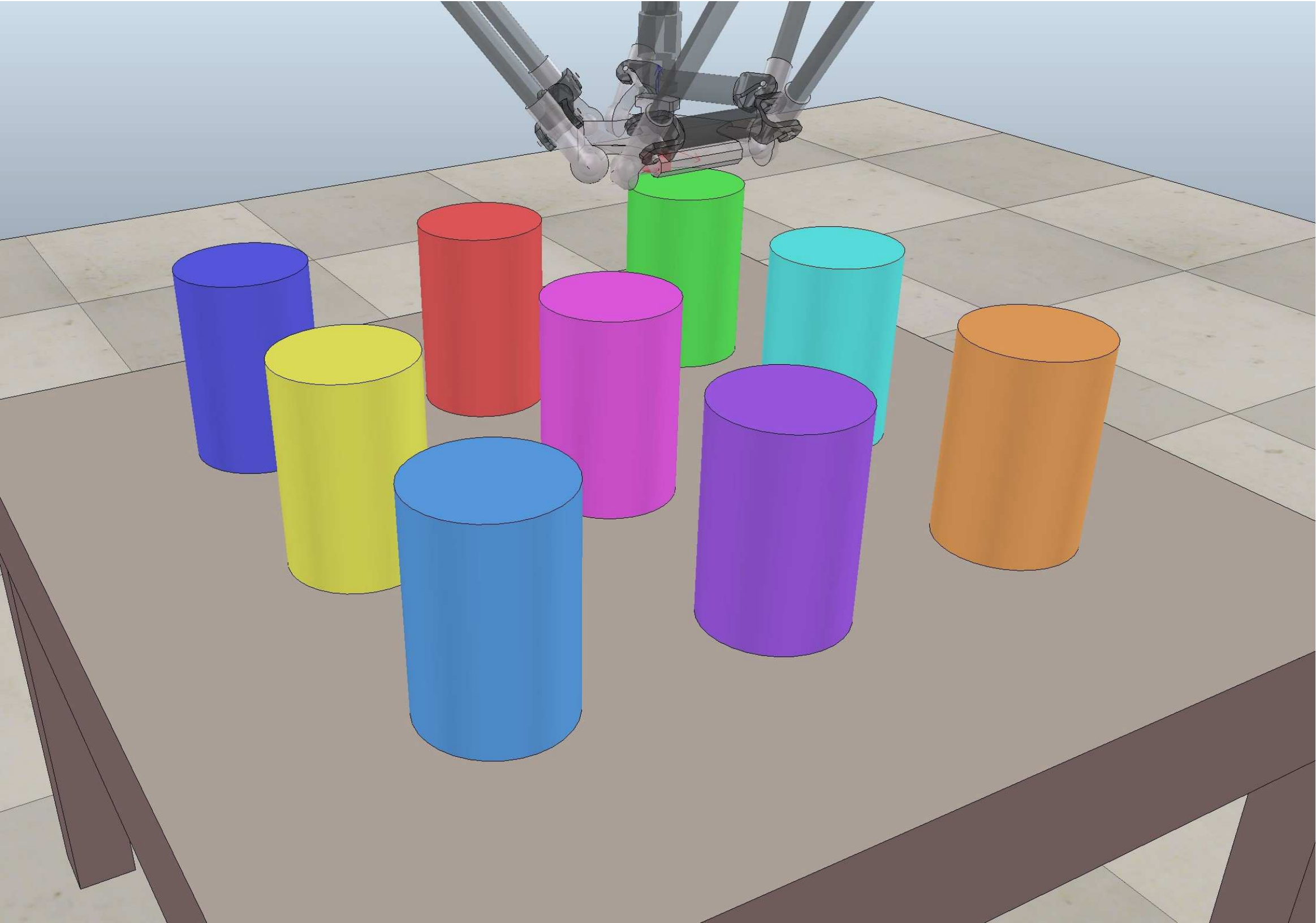} &
        \includegraphics[height=1.3in]{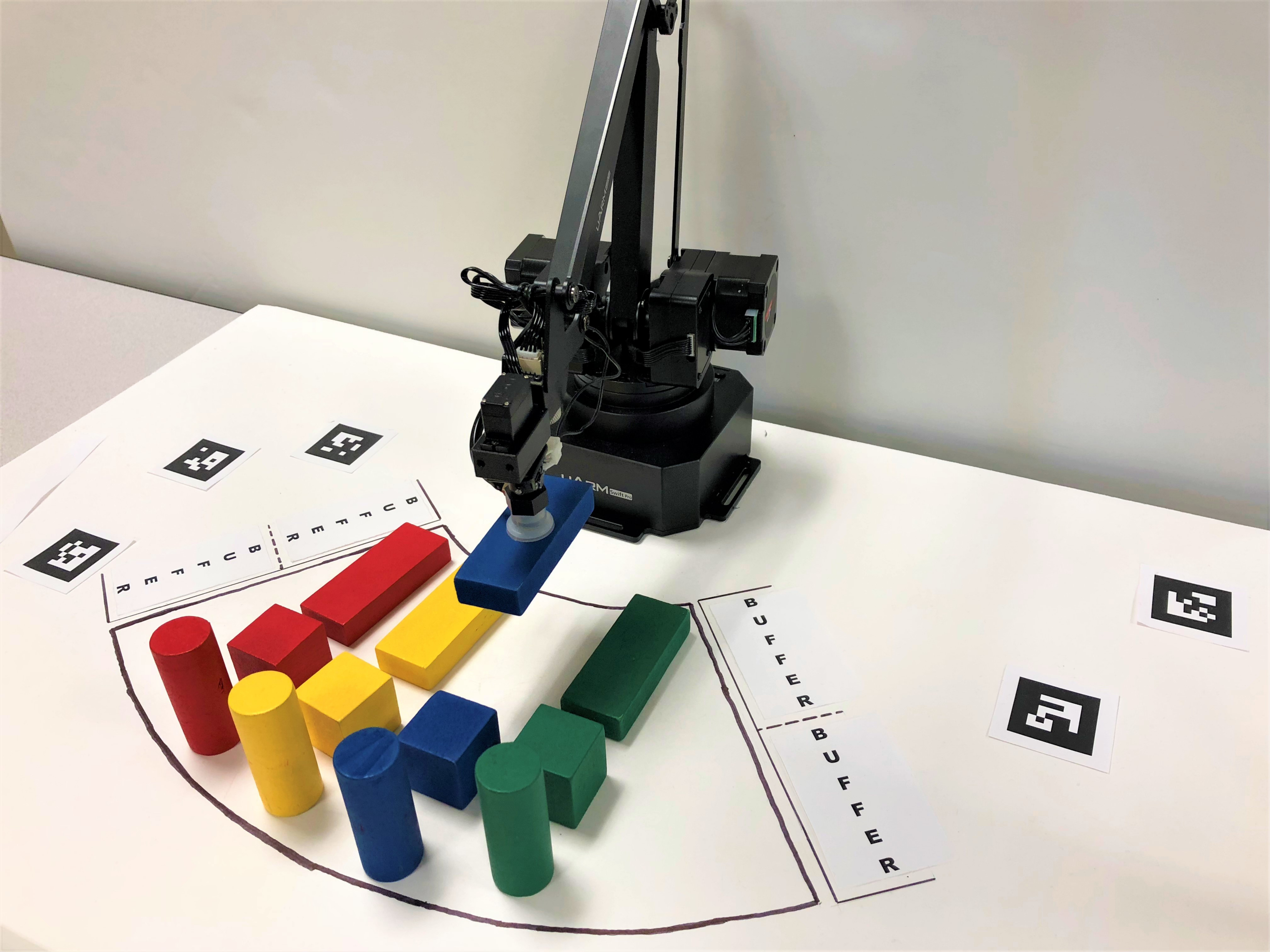} \\
        (a) & (b) & (c) & (d) \\    
    \end{tabular}
    \caption{%
        (a, b, c) An example of an object rearrangement challenge considered 
        in this work from a V-REP simulation~\citep{RohSinFre13}, 
        where the initial (b) and final (c) object poses are
        overlapping and an object needs to be placed at an intermediate
        location.
        (d) In addition to the V-REP simulations, the solutions
        presented in this work have been tested on an experimental hardware
        platform.%
    }
   \label{fig:example}
\end{figure*}

There are two important variations of this problem. The first requires
that the target object poses do not overlap with the initial ones.  In
this scenario, the number of pick-and-place actions is equal to the
number of objects not in their goal pose. Thus, the solution quality
is dependent upon the sequence with which the objects are transferred.
A good sequence can minimize the distance that the robot's
end-effector travels.  The second variant of the problem allows for
target poses to overlap with the initial poses, as in Fig.
\ref{fig:example}a-c. The situation sometimes necessitates the
identification of intermediate poses for some objects to complete the
task. In such cases, the quality of the solution tends to be dominated
by the number of intermediate poses needed to solve the problem, which
correlates to the number of pick-and-place actions the robot must
carry out. The primary objective is to find a solution, which uses the
minimum number of intermediate poses and among them minimize the
distance the robot's end-effector travels.

Both variations include some assumptions that simplify these instances
relative to the general rearrangement problem.  The non-overlapping
case in particular seems to be quite easy since a random feasible
solution can be trivially acquired. Nevertheless, this paper shows
that even in this simpler setup, the optimal variant of the problem
remains computationally hard. This is achieved by reducing the
Euclidean-{\tt TSP} problem \citep{Pap77} to the cost-optimal,
non-overlapping tabletop object rearrangement problem. Even in the
unlabeled case, where objects can occupy any target pose, the problem
is still hard. For overlapping initial and final poses, the paper
employs a graphical representation from the literature
\citep{BerSno+09}, which leads to the result that finding the minimum
number of pick-and-place actions relates to a well-known problem in
the algorithmic community, the ``Feedback Vertex Set'' ({\tt FVS})
problem \citep{Kar72}. This again indicates the hardness of the
challenge.

The benefit of these two-way reductions, beyond the hardness results
themselves, is that they suggest algorithmic solutions and provide an
expectation on the practical efficiency of the methods. In particular,
Euclidean-{\tt TSP} admits a polynomial-time approximation scheme ({\tt PTAS})
and good heuristics, which implies very good practical solutions for the
non-overlapping case.  On the other hand, the {\tt FVS} problem is {\tt
APX}-hard \citep{Kar72, DinSaf05}, which indicates that efficient algorithms are
harder for the overlapping case.  This motivated the consideration of
alternative heuristics for solving such challenges that make sense in the
context of object rearrangement.

The algorithms proposed here, which arose by mapping the object
rearrangement variations to well-studied problems have been evaluated
in terms of practical performance. For the non-overlapping case, an
alternative solver exists that was developed for a related challenge
\citep{TrePavFra13}. The {\tt TSP} solvers achieve superior
performance relative to this alternative when applied to object
rearrangement.  They achieve sub-second solution times for hundreds of
objects. Optimal solutions are shown to be significantly better than
the average, random, feasible solution.  For the overlapping case,
exact and heuristic solvers are considered.  The paper shows that
practically efficient methods achieve sub-second solution times
without a major impact in solution quality for tens of objects.

This article expands an earlier version of this work~\citep{HanSti+17}
and includes the following new contributions: {\em (i)} a proof
showing that cost-optimal unlabelled non-overlapping tabletop
rearrangement is NP-hard, {\em (ii)} a complete description of
ILP-based heuristics and an algorithm for solving the object
rearrangement problem with overlap sub-optimally, {\em (iii)}
significantly enhanced evaluation with experiments conducted on a
hardware platform (Fig.~\ref{fig:example}d), corroborating the
real-world benefits of the proposed method.

The structure of this manuscript is as follows.
Section~\ref{sec:related-work} reviews related work, followed by a
formal problem statement in Section~\ref{sec:preliminaries}.
Section~\ref{sec:rno} considers the object rearrangement problem when
the objects have non-overlapping start and goal arrangements.  The
more computationally challenging problem that involves overlapping
start and goal arrangements appears in Section~\ref{sec:rwo}.  The
evaluation of the proposed methods is broken down into two components:
simulation (Section~\ref{sec:simulations}) and physical experiments
(Section~\ref{sec:hardware-exp}).  The paper concludes with a summary
and remarks about future directions in Section~\ref{sec:conclusion}.

\section{Contribution Relative to Prior Work}\label{sec:related-work}
\emph{Multi-body planning} is a related challenge that is itself hard. In the
general, continuous case, complete approaches do not scale even though methods
exist that try to decrease the effective DOFs \citep{AroBer+99}. For specific
geometric setups, such as unlabeled unit-discs among polygonal obstacles,
optimality can be achieved \citep{SolYu+15}, even though the unlabeled case is
still hard \citep{SolHal15}. Given the hardness of multi-robot planning,
decoupled methods, such as priority-based schemes \citep{BerOve05} or velocity
tuning \citep{LerLauSim99}, trade completeness for efficiency. Assembly planning
\citep{WilLat94, HalLatWil00, SunRemAma01} deals with similar problems but few
optimality arguments have been made.

Recent progress has been achieved for the discrete variant of the problem,
where robots occupy vertices and move along edges of a graph. For this problem,
also known as ``pebble motion on a graph'' \citep{KorMilSpi84, CalDumPac08,
AulMon+99, GorHas10}, feasibility can be answered in linear time and paths can
be acquired in polynomial time. The optimal variation is still hard but recent
optimal solvers with good practical efficiency have been developed either by
extending heuristic search to the multi-robot case \citep{WagKanCho12,
SharSte+15}, or utilizing solvers for other hard problems, such as network-flow
\citep{YuLaV12a, YuLaV16}. The current work is motivated by this progress and
aims to show that for certain useful rearrangement setups it is possible to
come up with practically efficient algorithms through an understanding of the
problem's structure.

\emph{Navigation among Movable Obstacles} ({\tt NAMO}) is a related
computationally hard problem \citep{Wil91, CheHwa91, DemORoDem00, NieStaOve06},
where a robot moves and pushes objects.  A probabilistically complete solution
exists for this problem \citep{BerSti+08}.  {\tt NAMO} can be extended to
manipulation among movable obstacles ({\tt MAMO}) \citep{StiSch+07} and
rearrangement planning \citep{BenRiv98, Ota04}. Monotone instances for such
problems, where each obstacle may be moved at most once, are easier
\citep{StiSch+07}. Recent work has focused on ``non-monotone'' instances
\citep{HavOzb+14, SriFan+14, GarLozKae14, KroSho+14, KroBek15a, KroBek16}.
Rearrangement with overlaps considered in the current paper includes
``non-monotone'' instances although other aspects of the problem are relaxed.
In all these efforts, the focus is on feasibility and no solution quality
arguments have been provided.  Asymptotic optimality has been achieved for the
related ``minimum constraint removal'' path problem \citep{Hau14}, which,
however, does not consider negative object interactions.

The \emph{Pickup and Delivery Problem} ({\tt PDP}) \citep{BerCor+07,
BerCorLap10} is a well-studied problem in operations research that is similar
to tabletop object rearrangement, as long as the object geometries are ignored.
The {\tt PDP} models the pickup and delivery of goods between different parties
and can be viewed as a subclass of vehicle routing \citep{Lap92} or dispatching
\citep{ChrEil69}.  It is frequently specified over a graph embedded in the {\tt
2D} plane, where a subset of the vertices are pickup and delivery locations. A
{\tt PDP} in which pickup and delivery sites are not uniquely paired is also
known as the NP-hard swap problem \citep{GarJoh79, AniHas92}, for which a
2.5-optimal heuristic is known \citep{AniHas92}.  Many exact linear programming
algorithms and approximations are available \citep{BeuOudWas04, HofLok06,
GriHal+07} when pickup and delivery locations overlap, where pickup must happen
some time after delivery.  The stacker crane problem ({\tt SCP})
\citep{FreHecKim76, TrePavFra13} is a variation of {\tt PDP} of particular
relevance as it maps to the non-overlapping case of labeled object
rearrangement.  An asymptotically optimal solution for {\tt SCP}
\citep{TrePavFra13} is used as a comparison point in the evaluation section.

This work does not deal with other aspects of rearrangement, such as arm motion
\citep{SimLau+04, BerSriKuf12, CohChiLik13, ZucRat+13} or grasp planning
\citep{CioAll09, BohMor+14}.  Non-prehensile actions, such as pushing, are also
not considered \citep{CosHer+11, DogSri11}. Similar combinatorial issues to the
ones studied here are also studied by integrated task and motion planners, for
most of which there are no optimality guarantees \citep{CamAlaGra09, PlaHag10,
SriFan+14, GarLozKae14, GhaLalAla15, DanKinCha+16}.  Recent work on
asymptotically optimal task planning is at this point prohibitively expensive
for practical use \citep{VegRoy16}.

\section{Problem Statement}\label{sec:preliminaries}
	
This section formally defines the considered challenges.
	
\subsection{Tabletop Object Rearrangement with Overhand Grasps}

Consider a workspace $\Wspace$ with static obstacles and a set of $n$
movable objects $\objects = \{o_1, \dots,o_n\}$. For $o_i \in
\objects$, $\Cspace_i$ denotes its configuration space. Then,
$\Fspace_i \subseteq \Cspace_i$ is the set of collision-free
configurations of $o_i$ with respect to the static obstacles in
$\Wspace$. An {\em arrangement} $R = \{r_1, \dots, r_n\}$ for the
objects $\objects$ specifies the configurations $r_i \in \Cspace_i$
for each object $o_i$.  A feasible arrangement is one satisfying:
\begin{enumerate}
\item $\forall\ r_i \in R, r_i \in \Fspace_i$;
\item $\forall\ r_i, r_j \in R$, if $i \neq j$, then objects $o_i$ and
  $o_j$ are not in collision when placed at $r_i$ and $r_j$,
  respectively.
\end{enumerate}

This work focuses on closed and bounded planar workspaces: $\mathcal W \subset
\mathbb R^2$. The setting is frequently referred to as the {\em
  tabletop} setup, in which the vertical projections of the objects on
the tabletop do not intersect. This work assumes that the manipulator
is able to employ {\em overhand} grasps, where an object can be
transferred after being lifted above all other objects.  In
particular, a pick-and-place operation of the manipulator involves
four steps:
\begin{enumerate}[\hspace*{1em}a.]
    \item bringing the end-effector above the object,
    \item grasping and lifting the object,
    \item transfer of the
        grasped object horizontally to its target (horizontal) location, and
    \item a downward motion prior to releasing the object. 
\end{enumerate}
This sequence constitutes a {\em manipulation action}.

The manipulator is initially at a rest position $s_M$ prior to
executing any pick-and-place actions and transitions to a rest
position $g_M$ at the conclusion of the rearrangement task. A 
{\em rest position} is a safe arm configuration, where there is no
collision with objects.

The illustrations that appear throughout the paper assume objects with
identical geometry. Nevertheless, the results derived in this paper
are not dependent on this assumption, i.e., objects need only be
general cylinders.\footnote{From differential 
geometry, a cylinder is defined as any ruled surface spanned by a
one-parameter family of parallel lines.}

Given the setup, the problem studied in the paper can be summarized as:
\begin{problem}
\textbf{Tabletop Object Rearrangement with Overhand grasps (\toro).}
Given feasible start and goal arrangements $R_S, R_G$ for objects
$\objects = \{o_1, \dots,o_n\}$ on a tabletop, determine a sequence of
collision-free pick-and-place actions with overhand grasps $\actions =
(a^1,a^2, \dots)$ that transfer $\objects$ from $R_S$ to $R_G$.
\end{problem}

A rearrangement problem is said to be {\em labeled} if objects are
unique and not interchangeable. Otherwise, the problem is {\em
  unlabeled}. If for two arbitrary arrangements $s \in R_S$ and $g \in
R_G$, the objects placed in $s$ and $g$ are not in collision, then the
problem is said to have {\em no overlaps}.  Otherwise, the problem is
said to have {\em overlaps}.

This paper primarily focuses on the labeled \toro case and identifies
an important subcase:
\begin{itemize}
  \item \textit{\toro with NO overlaps (\rno)}
\end{itemize}

\remark The partition of Problem 1 into the general \rwo case and the
subcase of \rno is not arbitrary. \rwo is structurally richer and
harder from a computational perspective.  Both versions of the problem
can be extended to the unlabeled and partially labeled variants.  This
paper does not treat the labeled and unlabeled variants as separate
cases but will briefly discuss differences that arise due to
formulation when appropriate.

\subsection{Optimization Criteria}
Recall that a manipulation action $a^i$ has four components: an initial
move, a grasp, a transport phase, and a release. Since grasping is
frequently the source of difficulty in object manipulation tasks, it
is assumed in the paper that grasps and subsequent releases induce the 
most cost in manipulation actions. The other source of cost can be 
attributed to the length of the manipulator's path. This part of the 
cost is captured through the Euclidean distance traveled by the end 
effector between grasps and releases. For a manipulation action $a^i$, 
the incurred cost is
\begin{equation}\label{equ:action}
c_{a^i} = c_md^i_e + c_g + c_md^i_l + c_r,
\end{equation}
where $c_m,~c_g,~c_r$ are costs associated with moving the
manipulator, a single grasp, and a single release,
respectively. $d_e^i$ and $d_l^i$ are the straight line distances
traveled by the end effector in the first (object-free) and third
(carrying an object) stages of a manipulation action, respectively.
		
The total cost associated with solving a \toro instance is then
captured by
\begin{equation}\label{equ:totalcost}
c_{T} = \sum_{i = 1}^{|\actions|} c_{a^i} = |\actions| (c_g + c_r) + c_m\Big(\sum_{i = 1}^{|\actions|} (d^i_e + d^i_l) + d_f\Big),
\end{equation}
\noindent where $d_f$ is the distance between the location of the last
release of the end effector and its rest position $g_M$. Of the two
additive terms in~\eqref{equ:totalcost}, note that the first term
dominates the second.  Because the absolute value of $c_g, c_r$, and
$c_m$ are different for different systems, the assignment of their
absolute values is left to practitioners. The focus of this paper is
the analysis and minimization of the two additive terms
in~\eqref{equ:totalcost}.

\subsection{Object Buffer Locations}
The resolution of \rwo (Section~\ref{sec:rwo}) may require the
temporary placement of some object(s) at intermediate locations
outside those in $R_S \cup R_G$. When this occurs, external buffer
locations may be used as temporary locations for object
placement. More formally, there exists a set of configurations $B =
\{b_1, b_2, \dots\}$, called \textsl{buffers}, which are available to
the manipulator and do not overlap with object placements in $R_S$ or
$R_G$. 

\remark This work, which focuses on the combinatorial aspects of
multi-object manipulation and rearrangement, utilizes exclusively
buffers that are not on the tabletop. It is understood that the number
of external buffers \textit{may} be reduced by attempting to first
search for potential buffers within the tabletop. Nevertheless, there
are scenarios where the use of external buffers may be necessary.

\section{TORO with No Overlaps (\rno)}\label{sec:rno}
When there is no overlap between any pair of start and goal configurations, 
an object can be transferred directly from its start configuration to its
goal configuration. A direct implication is that an optimal sequence of 
manipulation actions contains exactly $|\actions| = |\objects| = n$ grasps and 
the same number of releases. Note that a minimum of $n$ grasps and 
releases are necessary. This also implies that no buffer is required 
since using buffers will incur additional grasp and release costs. 
Therefore, for \rno,~\eqref{equ:totalcost} becomes 
\begin{equation}\label{equ:tcno}
c_{T} = n(c_g + c_r) + c_m\Big(\sum_{i = 1}^{n} (d^i_e + d^i_l) + d_f\Big),
\end{equation}
i.e., only the distance traveled by the end effector affects the cost. 
The problem instance that minimizes~\eqref{equ:tcno} is referred to as Cost-optimal \rno.
The following theorem provides a hardness result for Cost-optimal \rno. 

\begin{theorem}\label{t:lno-np-hard}Cost-optimal \rno\ is NP-hard.  
\end{theorem} 
\noindent\begin{IEEEproof}
Reduction from Euclidean-\tsp\ \citep{Pap77}. Let $p_0,p_1,\ldots,
p_n$ be an arbitrary set of $n+1$ points in 2D. The set of points 
induces an Euclidean-\tsp. Let $d_{ij}$ denote the Euclidean distance 
between $p_i$ and $p_j$ for $0 \le i, j \le n$. In the formulation 
given in \citep{Pap77}, it is assumed that $d_{ij}$ are integers, which 
is equivalent to assuming the distances are rational numbers. To reduce 
the stated \tsp problem to a cost-optimal \rno\ problem, pick some 
positive $\varepsilon \ll 1/(4n)$. Let $p_0$ be the rest position of 
the manipulator in an object rearrangement problem. For each $p_i$, 
$1 \le i \le n$, split $p_i$ into a pair of start and goal configurations 
$(s_i, g_i)$ such that {\it (i)} $p_i = \frac{s_i + g_i}{2}$, {\it (ii)} 
$s_{i2} = g_{i2}$, and {\it (iii)} $s_{i1} + \varepsilon = g_{i1}$. An 
illustration of the reduction is provided in Fig.~\ref{fig:rno-reduction}. 
The reduced \rno instance is fully defined by $p_0$, $R_S = \{s_1, \ldots,
s_n\}$ and $R_G = \{g_1, \ldots, g_n\}$. A cost-optimal (as defined 
by~\eqref{equ:tcno}) solution to this \rno\ problem induces a (closed) 
path starting from $p_0$, going through each $s_i$ and $g_i$ exactly once, 
and ending at $p_0$. Moreover, each $g_i$ is visited immediately
after the corresponding $s_i$ is visited. Based on this path, the manipulator
moves to a start location to pick up an object, drop the object at the
corresponding goal configuration, and then move to the next object until all
objects are rearranged. Denote the loop path as $P$ and let its total length be 
$D$.

\begin{figure}[htp]
	\begin{center}
	\begin{tabular}{ccc}
	\begin{tikzpicture}[scale=1]
		\foreach \nodeName/\nodeLocation in {p_0/{(0, 2)}, p_1/{(-1, 1)}, p_2/{(1, 0)}}{
			\node (\nodeName) at \nodeLocation {$\nodeName$};
		}
		\foreach \edgeFrom/\edgeTo in {p_0/p_1, p_1/p_2, p_2/p_0}{
			\draw [-] (\edgeFrom) to (\edgeTo);
		}
	\end{tikzpicture}
	&&
	\begin{tikzpicture}[scale=1]
		\foreach \nodeName/\nodeLocation in {p_0/{(0, 2)}, s_1/{(-1.5, 1)}, 
		g_1/{(-0.5, 1)}, s_2/{(0.5, 0)}, g_2/{(1.5, 0)}}{
			\node (\nodeName) at \nodeLocation {$\nodeName$};
		}
		\foreach \edgeFrom/\edgeTo in {p_0/s_1, s_1/g_1, g_1/s_2, s_2/g_2, g_2/p_0}{
			\draw [-] (\edgeFrom) to (\edgeTo);
		}
		\node[draw=none, fill=none] at (1, 0.15) {$\varepsilon$};
		\node[draw=none, fill=none] at (-1, 1.15) {$\varepsilon$};
	\end{tikzpicture}\\
	{\footnotesize (a)} && {\footnotesize (b)}\\
	\end{tabular}
	\end{center}
	\caption{\label{fig:rno-reduction} Reduction from Euclidean-\tsp{} to cost-optimal \rno{}}  
\end{figure}
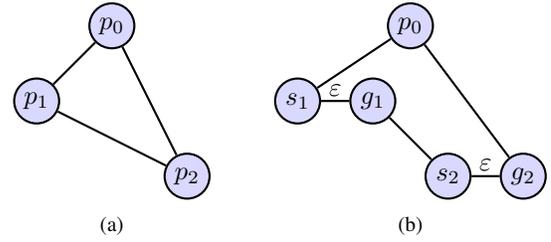

Assume that the Euclidean-\tsp has an optimal solution path $P_{opt}$ 
with a total distance of $D_{opt}$ (an integer). Then $P$ from solving the cost-optimal 
\rno\ yields such an optimal path for the \tsp. To show this, from $P$, 
simply contract the edges $s_ig_i$ for all $1 \le i \le n$. This clearly 
yields a solution to the Euclidean-\tsp; let the resulting path be $P'$ 
with total length $D'$. As edges are contracted along $P$, by the triangle 
inequality, $D' \le D$. It remains to show that $D' = D_{opt}$. Suppose this 
is not the case, then $D' \ge D_{opt} + 1$. However, if this is the case, 
a solution to the \rno can be constructed by splitting $p_i$ into $s_i$ and $g_i$ 
along $P_{opt}$. It is straightforward to establish that the total distance
of this \rno path is bounded by $D_{opt} + n\varepsilon < D_{opt} 
+ n*1/(4n) = D_{opt} + 1/4 < D_{opt} + 1 \le D' \le D$. Since this is a 
contradiction, $D' = D_{opt}$. 
\end{IEEEproof}

\remark Note that an NP-hardness proof of a similar problem can be
found in \citep{FreGua93}, as is mentioned in
\citep{TrePavFra13}. Nevertheless, the problem is stated for a tree and
is non-Euclidean. Furthermore, it is straightforward to show that the
decision version of the cost-optimal \rno\ problem is NP-complete;
the detail, which is non-essential to the focus of the current paper, 
is omitted.

\remark Interestingly, \rno\ may also be reduced to a variant of \tsp 
with very little overhead. Because highly efficient \tsp solvers are 
available, the reduction route provides an effective approach for 
solving \rno{}. That is, an \rno\ instance may be reduced to \tsp and
solved, with the \tsp solution readily translated back to a solution 
to the \rno instance that is cost-optimal. This is not always a feature 
of NP-hardness reductions. The straightforward algorithm for the 
computation is outlined in Alg.~\ref{algo:lno}. The inputs to the 
algorithm are the rest positions of the manipulator and the start and 
goal configurations of objects. The output is the solution for \rno, 
represented as a sequence of manipulation actions $\actions$, which 
has completeness and optimality guarantees.

\begin{algorithm}
	\small
	\DontPrintSemicolon
	\KwIn{Configurations $s_M$, $g_M$, Arrangements $R_S$, $R_G$.}
	\KwOut{A sequence of manipulation actions $\actions$.}	
	$G_{NO} \leftarrow ${\sc ConstructTSPGraph}$(R_S, R_G, s_M, s_G)$\;\label{algo:lno_model}
	$S_{raw}  \gets$ {\sc SolveTSP}$(G_{NO})$\;\label{algo:lno_solve}
	$\actions \gets$ {\sc RetrieveActions}$(S_{raw})$\;\label{algo:rno_action}
	\Return{$\actions$}\;
	\caption{{\sc ToroNoTSP}}
	\label{algo:lno}
\end{algorithm}
	 
\begin{figure}[htp]
	\centering
	\begin{tikzpicture}[scale=1]
		\foreach \nodeName/\nodeLocation in {s_M/{(0, 0)}, s_1/{(1, 1)}, 
		s_2/{(1, 0)}, u_1/{(2, 1)}, u_2/{(2, 0)}, g_1/{(3, 1)}, 
		g_2/{(3, 0)}, g_M/{(4, 0)}, u_0/{(2, -1)}}{
			\node (\nodeName) at \nodeLocation {$\nodeName$};
		}
		\foreach \edgeFrom/\edgeTo in {s_M/s_1, s_1/u_1, u_1/g_1, g_1/g_M, 
		s_M/s_2, s_2/u_2, u_2/g_2, g_2/g_M, s_M/u_0, u_0/g_M, s_1/g_2, s_2/g_1}{
			\draw [-] (\edgeFrom) to (\edgeTo);
		}
	\end{tikzpicture}
	\caption{An example of $G_{NO}$ for 2 objects. The nodes 
	$s_M$ and $g_M$ denote the initial and final rest positions of the 
	manipulator end effector.}
	\label{fig:rno-tsp}
\end{figure}
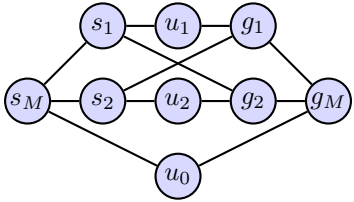

At Line \ref{algo:lno_model} of Alg.~\ref{algo:lno}, a graph
$G_{NO}(V_{NO}, E_{NO})$ is generated as the input to the \tsp
problem.  The graph is constructed from the \rno\ instance as
follows. A vertex is created for each element of $R_S$ and
$R_G$. Then, a complete bipartite graph is created between these two
sets of vertices. A set of vertices $U = \{u_1, \ldots, u_{|R_S|}\}$
is then inserted into edges $s_ig_i$ for $1 \le i \le
|R_S|$. Afterward, $s_M$ (resp., $g_M$) is added as a vertex and is
connected to $s_i$ (resp., $g_i$) for $1 \le i \le |R_S|$. Finally, a
vertex $u_0$ is added and connected to both $s_M$ and $g_M$. See
Fig.~\ref{fig:rno-tsp} for the straightforward example for $|R_S| =
2$.
	
Let $w(a, b)$ denote the weight of an edge $(a, b) \in E_{NO}$. For 
all $1 \leq i, j \leq n, i \neq j$ ($\text{dist}(x, y)$ denotes the	
Euclidean distance between $x$ and $y$ in 2D):
\begin{align*}
	& w(s_M, u_0) = w(g_M, u_0) = 0, 
	w(s_M, s_i) = \text{dist}(s_M, s_i), \\
	& w(g_M, g_i) = \text{dist}(g_M, g_i), 
	w(s_i, u_i) = w(u_i, g_i) = 0, \\
	& w(s_i, g_j) = \text{dist}(s_i, g_j).	
\end{align*}
With the construction, a \tsp tour through $G_{NO}$ must use
$s_Mu_0g_M$ and all $s_iu_ig_i$ for all $1 \le i \le |R_S|$. To form a
complete tour, exactly $(|R_S| - 1)$ edges of the form $g_is_j$, where
$i \ne j$ must be used. At Line \ref{algo:lno_solve}, the \tsp is
solved (using Concorde \tsp solver \citep{AppBix+07}). This yields a
minimum weight solution $S_{raw}$, which is a cycle containing all $v
\in V_{NO}$. The manipulation actions can then be retrieved (Line
\ref{algo:rno_action}).

An alternative solution to \rno\ could employ the asymptotically optimal, 
\textsl{SPLICE} algorithm, introduced in \citep{TrePavFra13}.

\subsection{Unlabelled \rno}

The scenario where objects are 
unlabeled is a special case of \rno which has significance in real-world applications (e.g., the 
pancake stacking application). This case is denoted as \uno\ (unlabeled, 
no overlap). Adapting the NP-hardness proof for the \rno\ problem 
shows that cost-optimal \uno\ is also NP-hard. Similar to the \rno\ 
case, the optimal solution only hinges on the distance traveled by the 
manipulator because no buffer is required and exactly $n$ grasps and 
releases are needed. 
\begin{theorem}\label{t:uno-np-hard}Cost-optimal \uno\ is NP-hard.  
\end{theorem} 
\begin{IEEEproof}
    See Appendix~\ref{proof-couno}.
\end{IEEEproof}

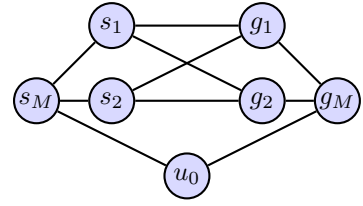
\begin{figure}[htp]
		\centering
		\begin{tikzpicture}[scale=1]
			\foreach \nodeName/\nodeLocation in {s_M/{(0, 0)}, s_1/{(1, 1)}, 
			s_2/{(1, 0)}, g_1/{(3, 1)}, g_2/{(3, 0)}, g_M/{(4, 0)}, u_0/{(2, -1)}}{
				\node (\nodeName) at \nodeLocation {$\nodeName$};
			}
			\foreach \edgeFrom/\edgeTo in {s_M/s_1, g_1/g_M, s_M/s_2, g_2/g_M, 
			s_M/u_0, u_0/g_M, s_1/g_2, s_2/g_1, s_1/g_1, s_2/g_2}{
				\draw [-] (\edgeFrom) to (\edgeTo);
			}
		\end{tikzpicture}
		\caption{An example of $G_{UNO}$ for 2 objects.}
		\label{fig:uno-tsp}
	\end{figure}

\vspace*{1em}
When solving a \uno instance, Alg. \ref{algo:lno} may be used with a 
few small changes. First, a different underlying graph must be constructed. 
Denote the new graph as $G_{UNO}(V_{UNO}, E_{UNO})$, where 
$V_{UNO} = R_S \cup R_G \cup \{s_M, u_0, g_M\}$. For all $1 \leq i,j \leq n$:
	\begin{align*}
		& w(s_M, u_0) = w(g_M, u_0) = 0,  w(s_M, s_i) = \text{dist}(s_M, s_i), \\
		& w(g_M, g_i) = \text{dist}(g_M, g_i), w(s_i, g_j) = \text{dist}(s_i, g_j).
	\end{align*}
All other edges are given infinite weight. An example of the updated 
structure of $G_{UNO}$ for two objects is illustrated in Fig.~\ref{fig:uno-tsp}. 
	
\section{TORO With Overlap (\rwo)}\label{sec:rwo}

Unlike \rno, \rwo\ has a more sophisticated structure and may require
buffers to solve. In this section, a {\em dependency graph}
\citep{BerSno+09} is used to model the structure of \rwo, which leads
to a classical NP-hard problem known as the {\em feedback vertex set}
problem \citep{Kar72}. The connection then leads to a complete
algorithm for optimally solving \rwo.

\subsection{The Dependency Graph and NP-Hardness of \rwo}

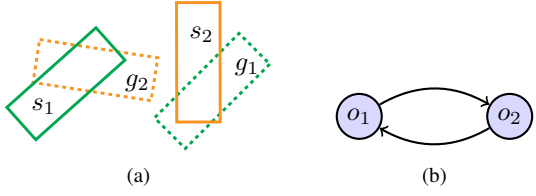
\begin{figure}
	\begin{center}
	\begin{tabular}{ccc}
  	\begin{overpic}[width=1.4in]{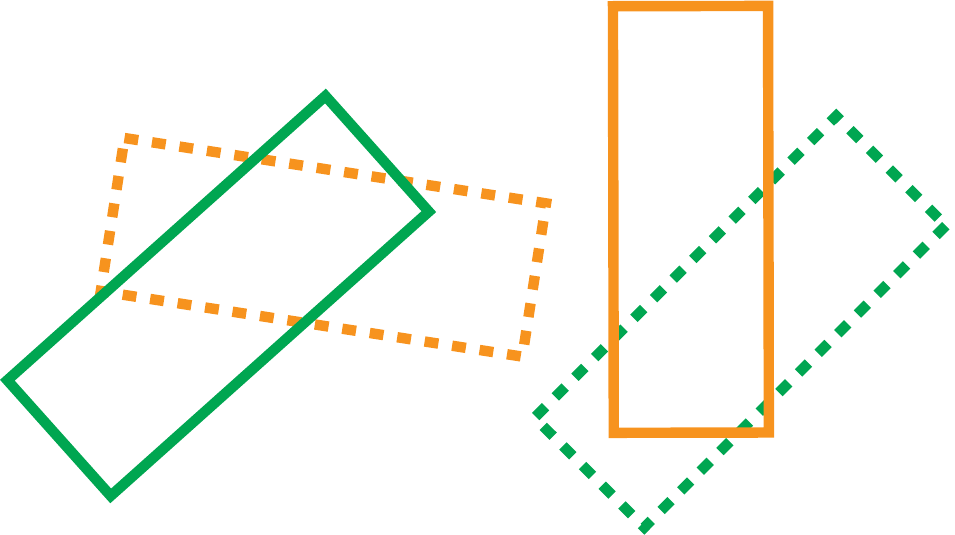}
	\put(10,15){{$s_1$}}
	\put(45,24){{$g_2$}}
	\put(69,42){{$s_2$}}
	\put(86,30){{$g_1$}}
	\end{overpic} &&
	
	\begin{tikzpicture}[scale=1]
		\node (o1) at (0, 0) {$o_1$};
		\node (o2) at (2, 0) {$o_2$};
		\draw [->] (o1) to [out = 30, in = 150] (o2);
		\draw [->] (o2) to [out = 210, in = -30] (o1);
	\end{tikzpicture}\\
			{\footnotesize (a)} && {\footnotesize (b)}\\
			\end{tabular}
			\end{center}
		\caption{Illustration of the dependency graph. (a) Two objects are to be moved 
		from $s_i$ to $g_i$, $i = 1, 2$. Due to the overlap between $s_1$ and $g_2$ as 
		well as the overlap between $s_2$ and $g_1$, one of the objects must be temporarily 
		moved aside. (b) The dependency graph capturing the scenario in (a).}
		\label{fig:dependencygraph}
	\end{figure}

Consider a {\em dependency digraph} $G_{dep}(V_{dep}, A_{dep})$, where 
$V_{dep} = \objects$, and $(o_i, o_j) \in A_{dep}$ \textit{iff} $g_i$ 
and $s_j$ overlap. Therefore, $o_j$ must be moved away from $s_j$ before 
moving $o_i$ to $g_i$. An example involving two objects is provided in 
Fig.~\ref{fig:dependencygraph}. The definition of dependency graph 
implies the following two observations. 
	
\begin{observation} If the out-degree of $o_i \in V_{dep}$ is 0, then $o_i$ 
can move to $g_i$ without collision. 
\end{observation}

\begin{observation}\label{l:lwo-dep-acyclic}
If $G_{dep}$ is not acyclic, solving \rwo\ requires at least $n + 1$ 
grasps.
\end{observation}

The dependency graph has obvious similarities to the well known {\em
feedback vertex set} (\fvs) problem \citep{Kar72}. A directed \fvs
problem is defined as follows. Given a strongly connected directed
graph $G = (V, A)$, an FVS is a set of vertices whose removal leaves
$G$ acyclic. Minimizing the cardinality of this set is NP-hard, even
when the maximum in degree or out degree is no more than two
\citep{GarJoh79}.  As it turns out, the set of removed vertices in an
\fvs problem mirrors the set of objects that must be moved to
temporary locations (i.e., buffers) for resolving the dependencies
between the objects, which corresponds to the additional grasps (and
releases) that must be performed in addition to the $n$ required
grasps for rearranging $n$ objects. The observation establishes that
cost-optimal \rwo\ is also computationally intractable. The following
lemma shows this point.
\begin{lemma}\label{l:lwo-min-grasp}
Let the dependency graph of a \rwo\ problem be a single strongly connected
graph. Then the minimum number of additional grasps required for solving the
\rwo\ problem equals the cardinality of the minimum FVS of the dependency 
graph. 
\end{lemma}
\begin{IEEEproof}
Given the dependency graph, let the additional grasps and releases be
$n_x$ and the minimum FVS have a cardinality of $n_{fvs}$, it remains
to show that $n_x = n_{fvs}$. First, if fewer than $n_{fvs}$ objects
are removed, which correspond to vertices of the dependency graph,
then there remains a directed cycle. By
Observation~\ref{l:lwo-dep-acyclic}, this part of the problem cannot
be solved. This establishes that $n_x \ge n_{fvs}$.  On the other
hand, once all objects corresponding to vertices in a minimum FVS are
moved to buffer locations, the dependency graph becomes acyclic. This
allows the remaining objects to be rearranged. This operation can be
carried out iteratively with objects whose corresponding vertices have
no incoming edges. On a directed acyclic graph (\DAG), there is always
such a vertex. Moreover, as such a vertex is removed from a \DAG, the
remaining graph must still be a \DAG and therefore must have either no
vertex (a trivial \DAG) or a vertex without incoming edges.
\end{IEEEproof}

For dependency graphs with multiple strongly connected components, the
required number of additional grasps and releases is simply the sum of
the required number of such actions for the individual strongly
connected components. 

For a fixed \rwo\ problem, let $n_{fvs}$ be the cardinality of the largest
(minimal) FVS computed over all strongly connected components of its dependency
graph. Then it is easy to see that the maximum number of required buffers is no
more than $n_{fvs}$. The NP-hardness of cost-optimal \rwo\ is established using
the reduction from \fvs problems to \rwo. This is more involved than reducing
\rwo\ to \fvs because the constructed \rwo\ must correspond to an actual
\toro problem in which the number of overlaps should not grow unbounded.

\begin{theorem}\label{t:lwo-min-buffer}
Cost-optimal \rwo is NP-hard. 
\end{theorem}
\begin{IEEEproof}
The \fvs problem on directed graphs is reduced to cost-optimal \rwo. An
\fvs problem is fully defined by specifying an arbitrary strongly
connected directed graph $G = (V, A)$ where each vertex has no more
than two incoming and two outgoing edges. A typical vertex
neighborhood can be represented as illustrated in
Fig.~\ref{fig:dep-temp}(a). Such a neighborhood is converted to a
dependency graph neighborhood of object rearrangement as follows. Each
of the original vertex $v_i \in V$ becomes an object $o_i$ which has
some $(s_i, g_i)$ pair as its start and goal configurations. For each
directed arc $v_iv_j$, split it into two arcs and add an
additional object $o_{ij}$. That is, create new arcs $o_io_{ij}$
and $o_{ij}o_j$ for each original arc $v_{ij}$ (see
Fig.~\ref{fig:dep-temp}(b)). This yields a dependency graph that is
again strongly connected. Two claims will be proven:
\begin{enumerate}
\item The constructed dependency graph corresponds to an object
  rearrangement problem, and
\item The minimum number of objects that must be moved away
  temporarily to solve the problem is the same as the size of the
  minimum FVS.
\end{enumerate}
\begin{figure}[htp]
	\begin{center}
	\begin{tabular}{ccc}
	\begin{tikzpicture}[scale=1.2]
		\foreach \nodeName/\nodeLocation in {v_1/{(0, 0)}, v_2/{(1, 1)}, v_3/{(0, 1)}, v_4/{(1, 0)}}{
			\node (\nodeName) at \nodeLocation {$\nodeName$};
		}
		\foreach \edgeFrom/\edgeTo in {v_1/v_4, v_3/v_1}{
			\draw [->] (\edgeFrom) to (\edgeTo);
		}
		\draw [->] (v_1) to [out = 30, in = -120] (v_2);
		\draw [->] (v_2) to [out = -150, in = 60] (v_1);
	\end{tikzpicture}
	&&
	\begin{tikzpicture}[scale=1]
		\foreach \nodeName/\nodeLocation in {o_1/{(0, 0)}, o_2/{(2, 2)}, o_3/{(0, 2)}, 
		o_4/{(2, 0)}, o_{31}/{(0, 1)}, o_{14}/{(1, 0)}, o_{12}/{(0.75, 1.25)}, o_{21}/{(1.25, 0.75)}}{
			\node (\nodeName) at \nodeLocation {$\nodeName$};
		}
		\foreach \edgeFrom/\edgeTo in {o_3/o_{31}, o_{31}/o_1, o_1/o_{12}, o_{12}/o_2,
		 o_2/o_{21}, o_{21}/o_1, o_1/o_{14}, o_{14}/o_4}{
			\draw [->] (\edgeFrom) to (\edgeTo);
		}
	\end{tikzpicture}\\
	{\footnotesize (a)} && {\footnotesize (b)}\\
	\end{tabular}
	\end{center}
	\caption{\label{fig:dep-temp} Converting a neighborhood of a graph 
	for an \fvs problem to parts of a dependency graph for a 
	\rwo problem.}  
\end{figure}
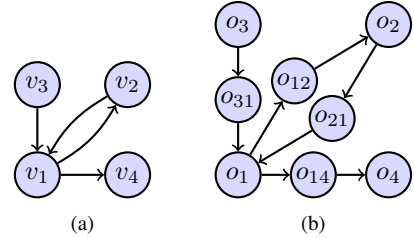
To prove the first claim, assume without loss of generality that the
objects have the same footprints on the tabletop.  Furthermore, only
the neighborhood of $o_1$ needs to be inspected because it is isolated
by the newly added objects. Recall that an incoming edge to $o_1$
means that the start configuration $o_1$ blocks the goals of some
other objects, in this case $o_{21}$ and $o_{31}$. This can be readily
realized by putting the goal configurations of $o_{21}$ and $o_{31}$
close to each other and have them overlap with the start configuration
of $o_1$. Note that the goal configurations of $o_{21}$ and $o_{31}$
have no other interactions. Therefore, such an arrangement is always
achievable for even simple (e.g., circular or square) footprints.
Similarly, for the outgoing edges from $o_1$, which mean other objects
block $o_1$'s goal, in this case $o_{12}$ and $o_{14}$, place the
start configurations of $o_{12}$ and $o_{14}$ close to each other and
make both overlap with the goal configuration of $o_1$. Again, the
start configurations of $o_{12}$ and $o_{14}$ have no other
interactions.

The second claim directly follows Lemma~\ref{l:lwo-min-grasp}. Now,
given an optimal solution to the reduced \rwo\ problem, it remains to
show that the solution can be converted to a solution to the original
\fvs problem. The solution to the \rwo\ problem provides a set of
objects that are moved to temporary locations. This yields a minimum
FVS on the dependency graph but not the original graph $G$. Note that
if a newly created object (e.g., $o_{ij}$) is moved to a temporary
place, either object $o_i$ or $o_j$ can be moved since this will
achieve no less in disconnecting the dependency graph. Doing this
across the dependency graph yields a minimum FVS for $G$.
\end{IEEEproof}
\remark It is possible to prove that \rwo\ is NP-hard using a similar
proof to the \rno\ case. To make the proof for
Theorem~\ref{t:lno-np-hard} work here, each $p_i$ can be split into an
overlapping pair of start and goal. Such a proof, however, would bury
the true complexity of \rwo, which is a much more difficult problem. Unlike the
Euclidean-\tsp problem, which admits $(1+\varepsilon)$-approximations
and good heuristics, \fvs problems are {\tt APX}-hard
\citep{Kar72,DinSaf05}.

\subsection{Algorithmic Solutions for \rwo}\label{sec:rwos}
\subsubsection{Feasible algorithm}
Once the link between a \rwo\ buffer requirement and \fvs is
established, an algorithm for solving \rwo\ becomes possible.  To do
this, an FVS set is found. Then the optimal rearrangement distance is
computed for this FVS set. The procedure for doing this is outlined in
\lwoalgs (Alg. \ref{algo:lwo}).  At Line \ref{algo:lwo-gdep}, the
dependency graph $G_{dep}$ is constructed.  At Line
\ref{algo:lwo-fvs1}-\ref{algo:lwo-fvs2}, an FVS is obtained for each
strongly connected component (SCC) in $G_{dep}$. Note that if these
FVSs are optimal, then the step yields the minimum number of required
grasps (and releases) as: $\min |\actions| = n + |B|.$

The residual work is to find the solution with $n + |B|$ grasps and the 
shortest travel distance (Line~\ref{algo:lwo-model}). The manipulation
actions
are then retrieved and returned.

\begin{algorithm}
	\small
	\DontPrintSemicolon
	\KwIn{Configurations $s_M$, $g_M$, Arrangements $R_S$, $R_G$}
	\KwOut{A set of manipulation actions $\actions$}	
	$G_{dep} \leftarrow ${\sc ConstructDepGraph}$(R_S, R_G)$\;\label{algo:lwo-gdep}
	$B \gets \emptyset$\;
	\For{each SCC \textbf{in} $G_{dep}$}{\label{algo:lwo-fvs1}
		$B \gets B\;\cup\;${\sc SolveFVS}$(\text{SCC})$\;\label{algo:lwo-fvs2}
	}
	$S_{raw} \gets ${\sc MinDist}$(s_M, g_M, R_S, R_G, G_{dep},B)$\; \label{algo:lwo-model}
  $\actions \gets ${\sc RetrieveActions}$(S_{raw})$\;
	\Return{$\actions$}\;
	\caption{{\sc ToroFVSSingle}}
	\label{algo:lwo}
\end{algorithm}

The paper explores two exact and three approximate methods as implementations
of {\sc SolveFVS}() (Line~\ref{algo:lwo-fvs2} of Alg.~\ref{algo:lwo}).  The two
exact methods are both based on integer linear programming (ILP) models,
similar to those introduced in \citep{BahSchNeu15}. They differ in how
cycle constraints are encoded: one uses a polynomial number of
constraints and the other simply enumerates all possible cycles.
Denote these two exact methods as {\bf ILP-Constraint} and {\bf ILP-Enumerate},
respectively. The details of these two exact methods are explained in 
Appendix~\ref{app:exact}. With regards to approximate solutions, several heuristic solutions are
presented:
\begin{enumerate}
\item {\bf Maximum Simple Cycle Heuristic (MSCH)}. The FVS is 
obtained by iteratively removing the node that appears on the most number of simple cycles in $G_{dep}$ until no more cycles exist. The simple cycles are 
enumerated.
\item {\bf Maximum Cycle Heuristic (MCH)}. This heuristic is similar to MSCH
but counts cycles differently. For each vertex $v \in V_{dep}$, it finds a 
cycle going through $v$ and marks the outgoing edge from $v$ on this cycle. 
The process is repeated for $v$ until no more cycles can be found. The vertex
with the largest cycle count is then removed first.
\item {\bf Maximum Degree Heuristics (MDH)}. This heuristic constructs an FVS
through vertex deletion based on the degree of the vertex until no cycles 
exist. 
\end{enumerate}
Based on FVS, the solution minimizing travel distance
can be found by {\sc MinDist}() (line \ref{algo:lwo-model}), which
is an LP modeling method inspired by \citep{YuLaV16} and
described in Appendix~\ref{app:lwolp}.

\subsubsection{Complete algorithm} Note that {\sc ToroFVSSingle}() is a 
complete algorithm for solving  \rwo\ but it is not a complete 
algorithm for solving \rwo\ optimally. With some additional engineering,
a complete optimal \rwo\ solver can also be constructed: under the 
assumption that grasping dominates the traveling costs, simply iterate
through all optimal FVS sets and then compute the subsequent minimum 
distance. After all such solutions are obtained, the optimal among these 
are chosen. It turns out that doing this enumeration does not 
provide much gain in solution quality as the optimal distances are very 
similar to each other. 

\section{Performance Evaluation of Simulations}\label{sec:simulations}

All simulations are executed on an Intel(R) Core(TM) i7-6900K CPU with 32GB RAM at
2133MHz. Concorde \citep{AppBix+07} is used for solving the \tsp and Gurobi
6.5.1 \citep{Gurobi} for ILP models. 

\subsection{TORO-NO: Minimizing the Travel Distance}
\begin{figure}[htp]
	\centering
	\includegraphics[keepaspectratio, width = .77\columnwidth]{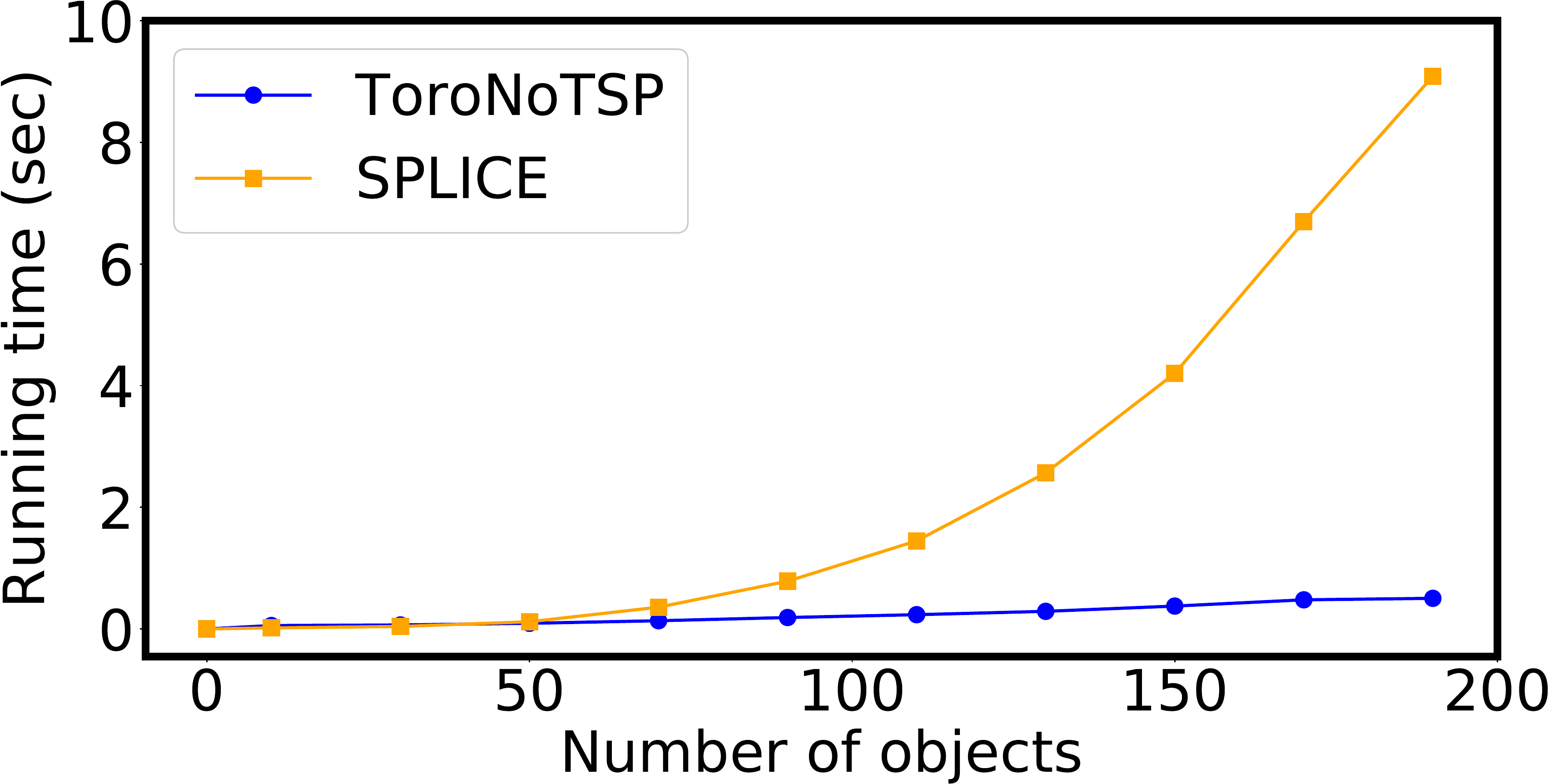}
	\caption{Running time comparison of \rnos and \textsl{SPLICE}.}
	\label{fig:rno-time}
\end{figure}

\begin{figure}[htp]
	\centering
	\includegraphics[keepaspectratio, width = .77\columnwidth]{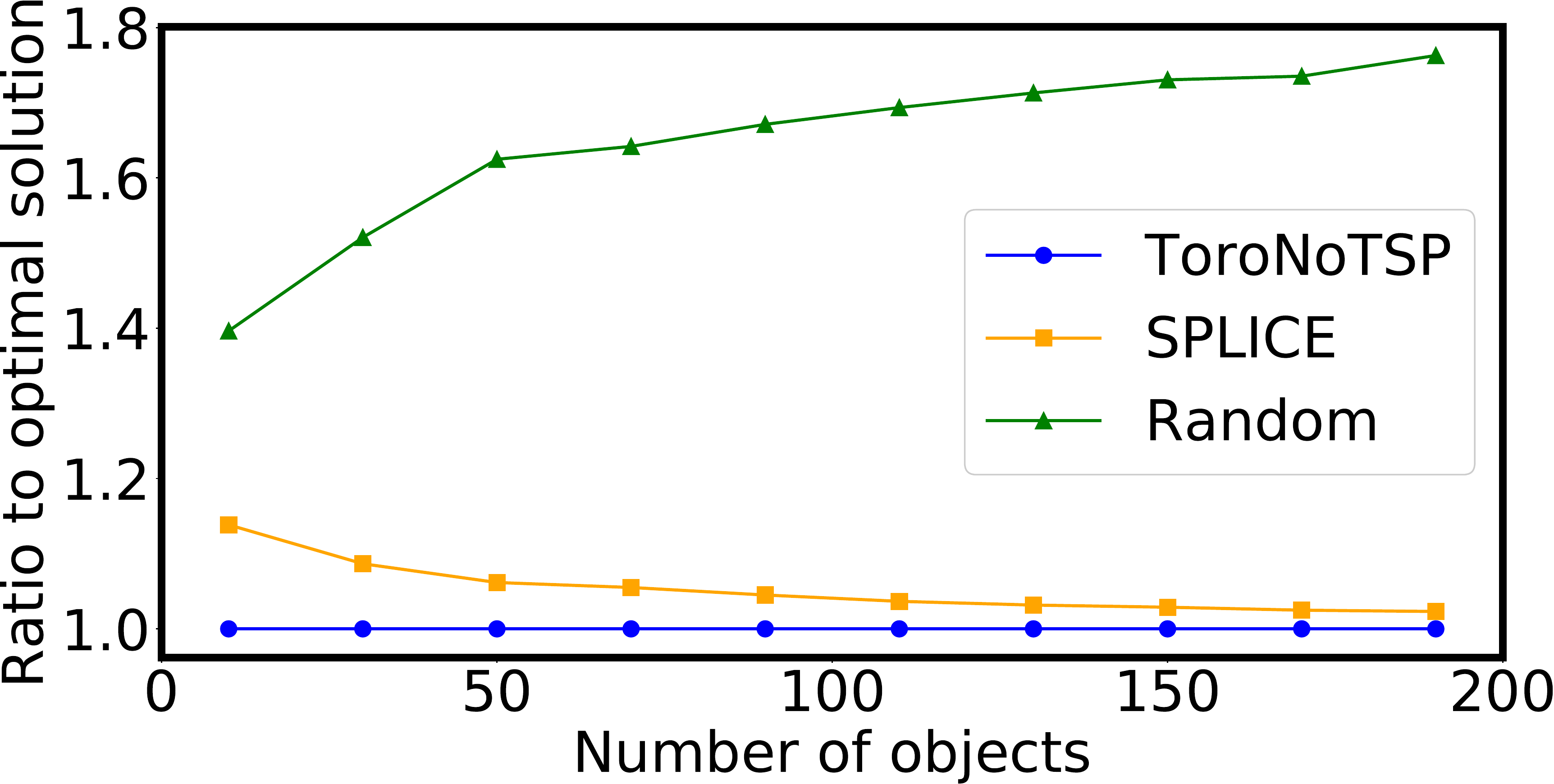}
	\caption{Optimality of \rnos, \textsl{SPLICE} and a random selection method.}
	\label{fig:rno-opt}
\end{figure}

To evaluate the effectiveness of \rnos, random \rno instances are generated 
in which the number of objects varies. For each choice of number of objects, 100
instances are tried and the average is taken. Although \rnos works on thousands 
of objects (it takes less than $30$ seconds for \rnos to solve instances
with $2500$ objects), the evaluation is limited to $200$
objects\footnote{State-of-the-art Delta robots have comparable
abilities.  For example, the Kawasaki YF03 Delta Robot is capable of performing
222 pick-and-place actions per minute (1kg objects).}.
Concerning running time, \rnos\ is compared with \textsl{SPLICE} \citep{TrePavFra13} 
which does not compute an exact optimal solution. As shown in 
Fig.~\ref{fig:rno-time}, it takes less than a second for \rnos to compute
the distance optimal manipulation action set. 
Fig.~\ref{fig:rno-opt} illustrates the solution quality of  \rnos, \textsl{SPLICE}, and an algorithm that 
picks a random feasible solution. Notice that the 
random feasible solution generally has poor quality. \textsl{SPLICE} does 
well as the number of objects increases, but under-performs compared to \rnos. 
In conclusion, \rnos provides the best performance on both running time
and optimality for practical sized \rno problems.  

For the unlabeled case (\uno), the same experiments are carried out. The results
appear in Table~\ref{tab:uno}. Note that \textsl{SPLICE} no longer applies.
The last line of the table is the optimality of random solutions, included 
for reference purposes. For larger cases, the \tsp based method is able to 
solve for over $500$ objects in $30$ seconds.
\begin{table}
	\small
	\centering
	\caption{Evaluation of the \tsp model for the unlabeled case.}
    \begin{tabularx}{\linewidth}{@{}|X| c | c | c | c|}
		\hline
		Number of objects & 10 & 50 & 100 & 200 \\ \hline
		Running time (sec) & 0.04 & 0.58 & 2.43 & 7.30 \\ \hline
		Optimality of random solution & 1.94 & 3.72 & 4.92 & 6.01 \\ \hline
	\end{tabularx}
	\label{tab:uno}
\end{table}

\subsection{TORO: Minimizing the Number of Grasps}

\begin{figure}[htp]
	    \centering
		\includegraphics[keepaspectratio, width = .77\columnwidth]{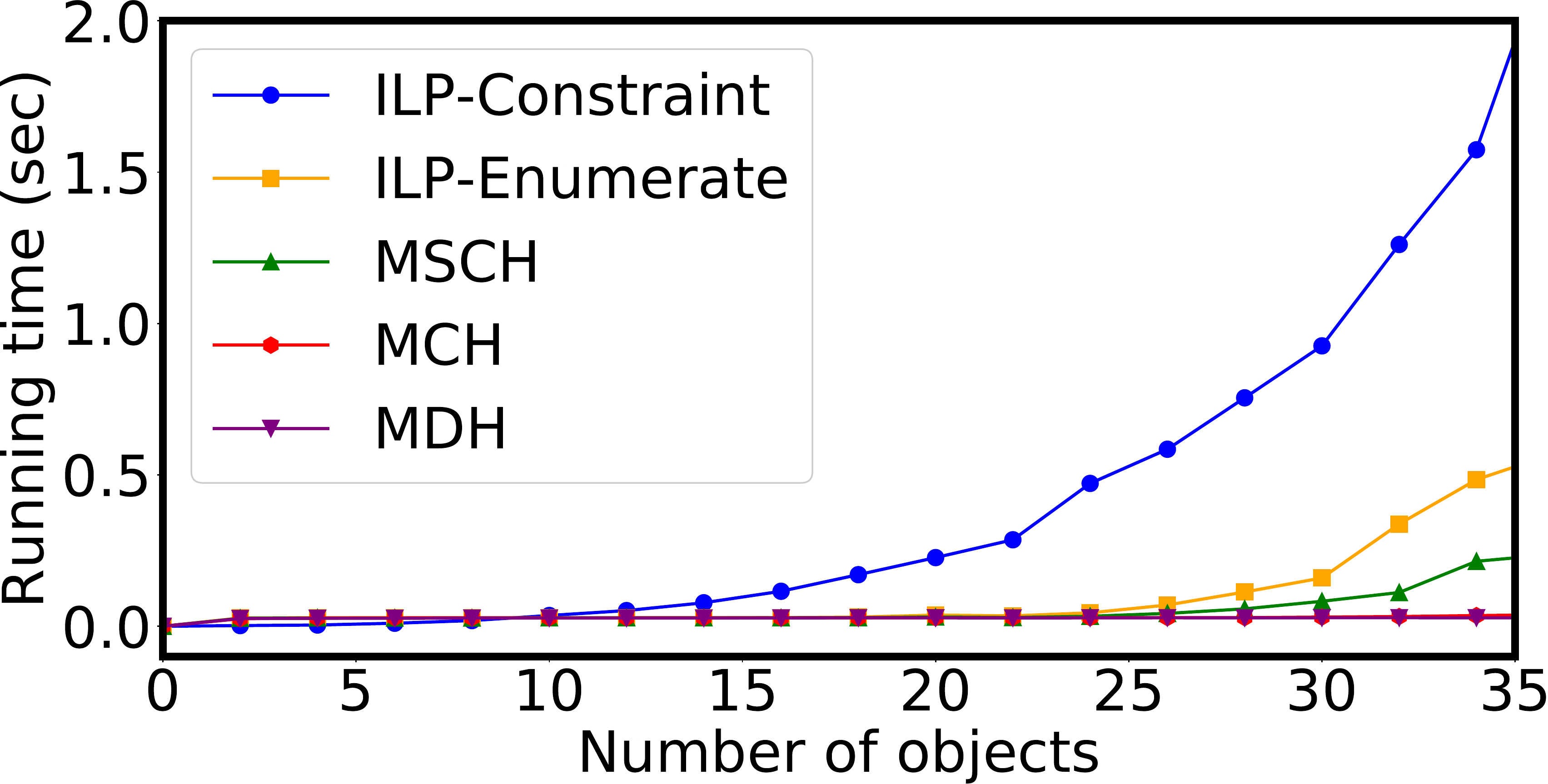}
		\caption{
            Running time of various methods for optimizing FVS. 
        }
		\label{fig:fvs-time}
	\end{figure}

\noindent To evaluate different FVS minimization methods, 
dependency graphs are generated by capping the average degree and maximum
degree for a fixed object count. To evaluate the running time, 
the average degree is set to $2$ and the maximum degree is set to $4$, which
creates significant dependencies.
The running time comparison is given in Fig.~\ref{fig:fvs-time} (averaged over
100 runs per data point). Although exact ILP-based methods took more time than
heuristics, they can solve optimally for over $30$ objects in just a few
seconds, which makes them very practical.

\begin{figure}[htp]
		\centering
		\includegraphics[keepaspectratio, width = .77\columnwidth]{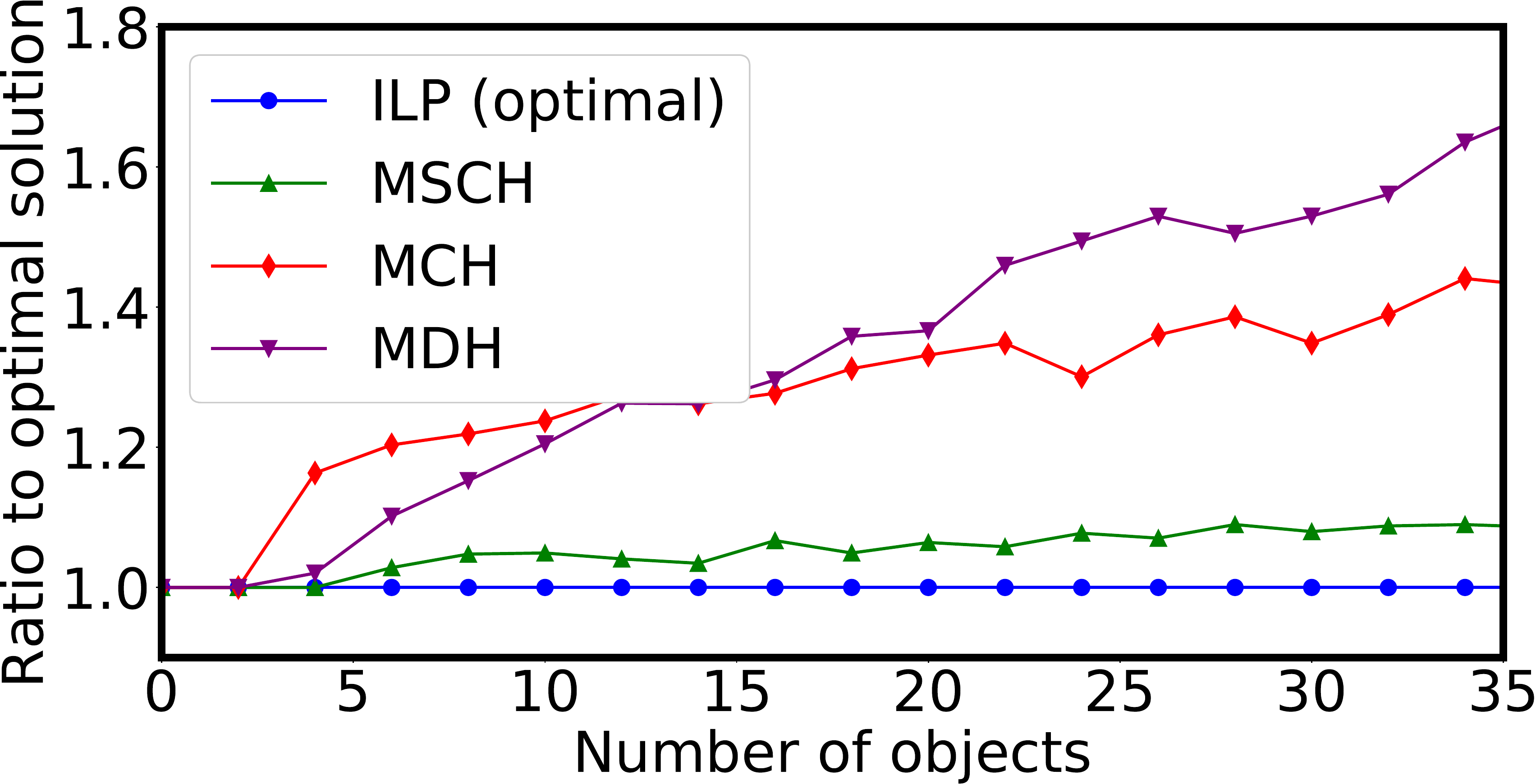}
        \caption{Optimality ratio of various methods for optimizing FVS as 
			compared with the optimal ILP-based methods.}
		\label{fig:fvs-opt}
	\end{figure}

When it comes to performance (Fig.~\ref{fig:fvs-opt}), ILP-based methods
have no competition. Interestingly, the simple cycle based method (MSCH) also
works quite well and may be useful in place of ILP-based methods for larger 
problems, given that MSCH runs faster. 
	
\begin{figure}[htp]
		\centering
		\includegraphics[keepaspectratio, width = .77\columnwidth]{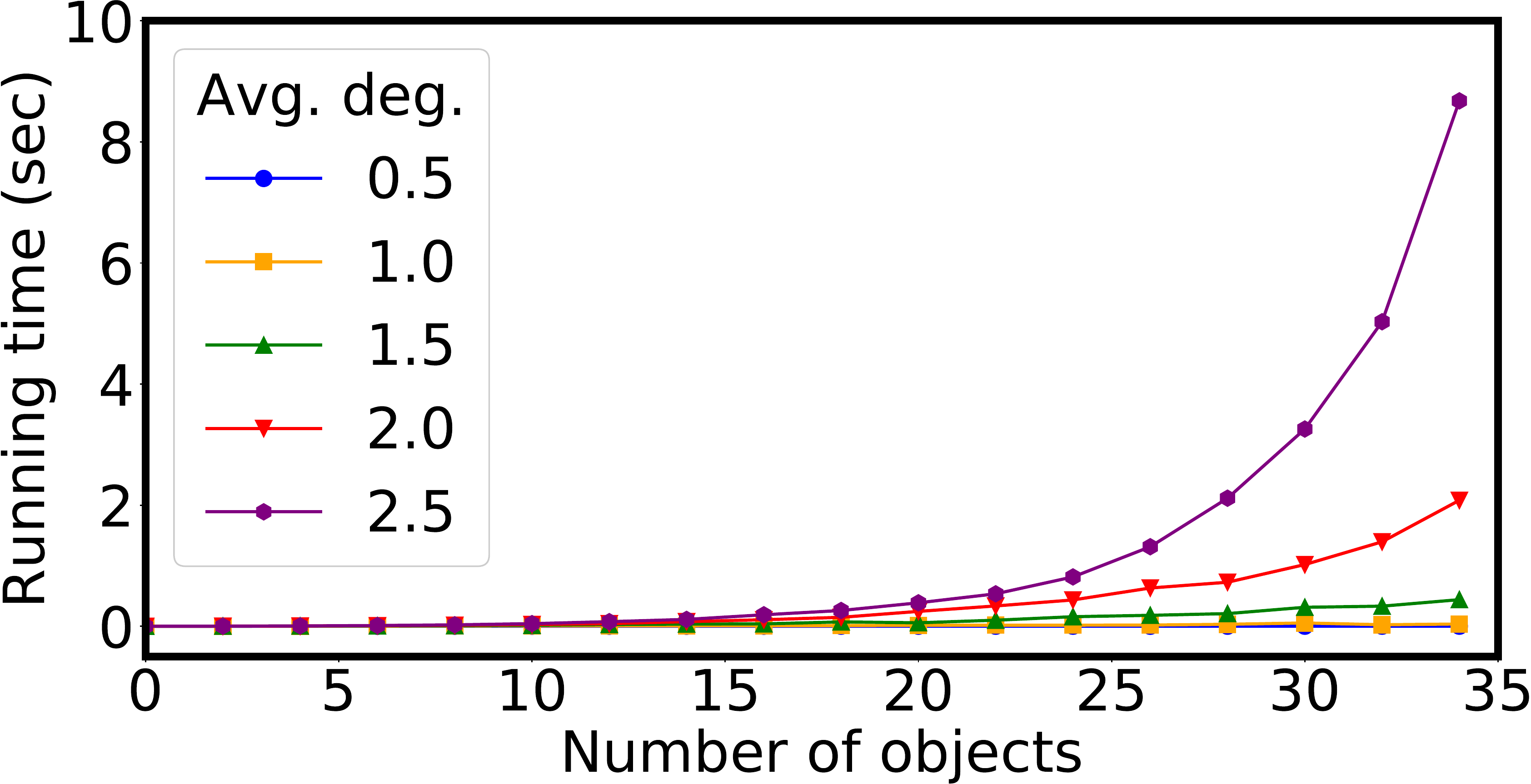}
		\caption{The running time of ILP-Constraint under varying $G_{dep}$
		average degree. Maximum degree is capped at twice the average degree.}
		\label{fig:fvs-dense-eval}
	\end{figure}

The performance is also affected by the average degree for each node,
which is directly linked to the complexity of $G_{dep}$. Fixating on
the ILP-Constraint algorithm, experiments with an average degree of
$0.5$-$2.5$ are included ($2.5$ average degree yields rather
constrained dependency graphs). As can be observed from
Fig.~\ref{fig:fvs-dense-eval}, for up to $35$ objects, an optimal FVS
can be readily computed in a few seconds.

\begin{figure}[htp]
		\centering
		\includegraphics[keepaspectratio, width = .77\columnwidth]{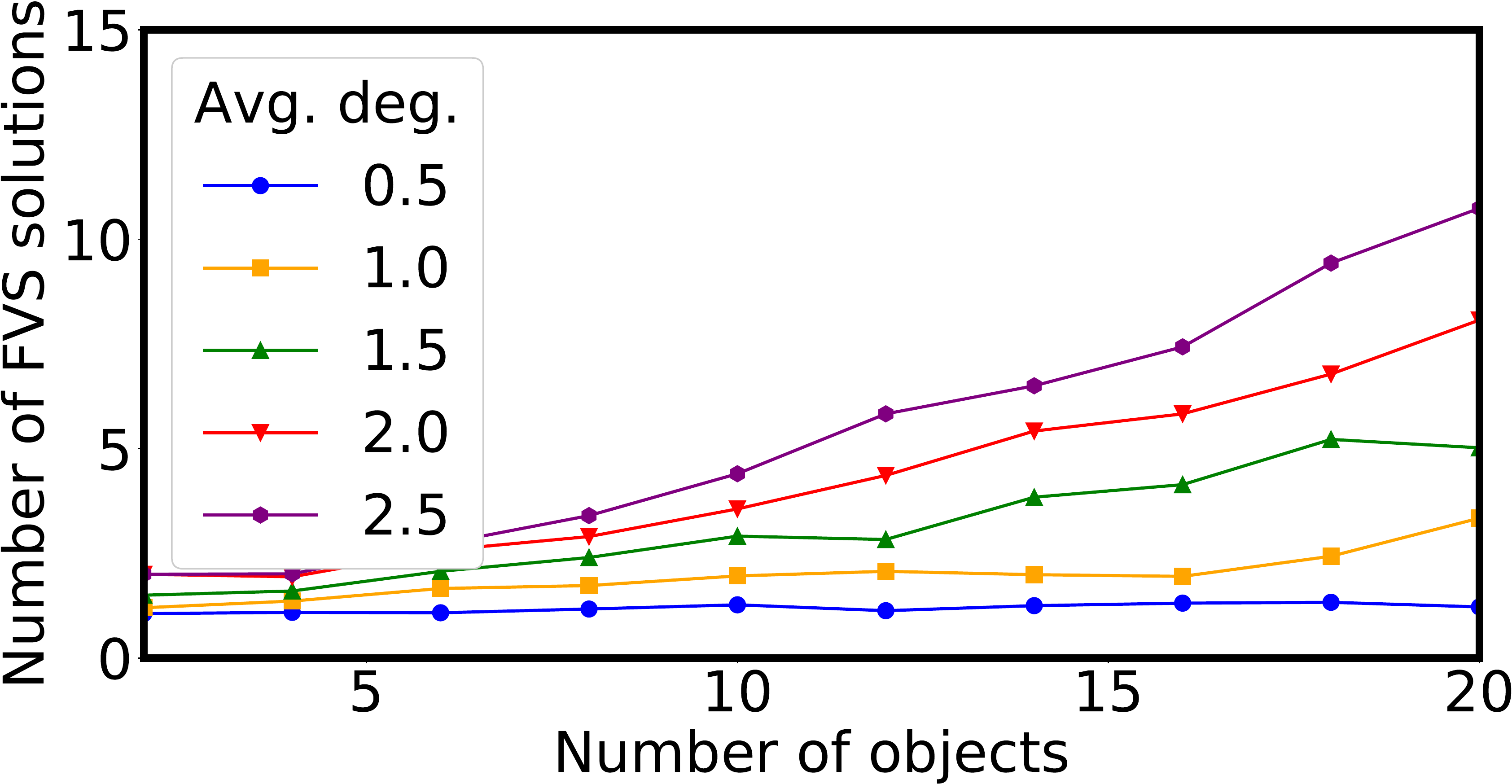}
		\caption{The number of optimal FVS solutions in expectation.}
		\label{fig:fvs-number}
	\end{figure}

Finally, this section emphasizes an observation regarding the number of optimal
FVS sets (Fig.~\ref{fig:fvs-number}). By disabling FVSs that are already
obtained in subsequent runs, all FVSs for a given problem can be exhaustively
enumerated for varying numbers of objects and average degree of $G_{dep}$. The
number of optimal FVSs turns out to be fairly limited.

\subsection{TORO: Overall Performance}
\begin{figure}[htp]
	\centering
	\includegraphics[keepaspectratio, width = .77\columnwidth]{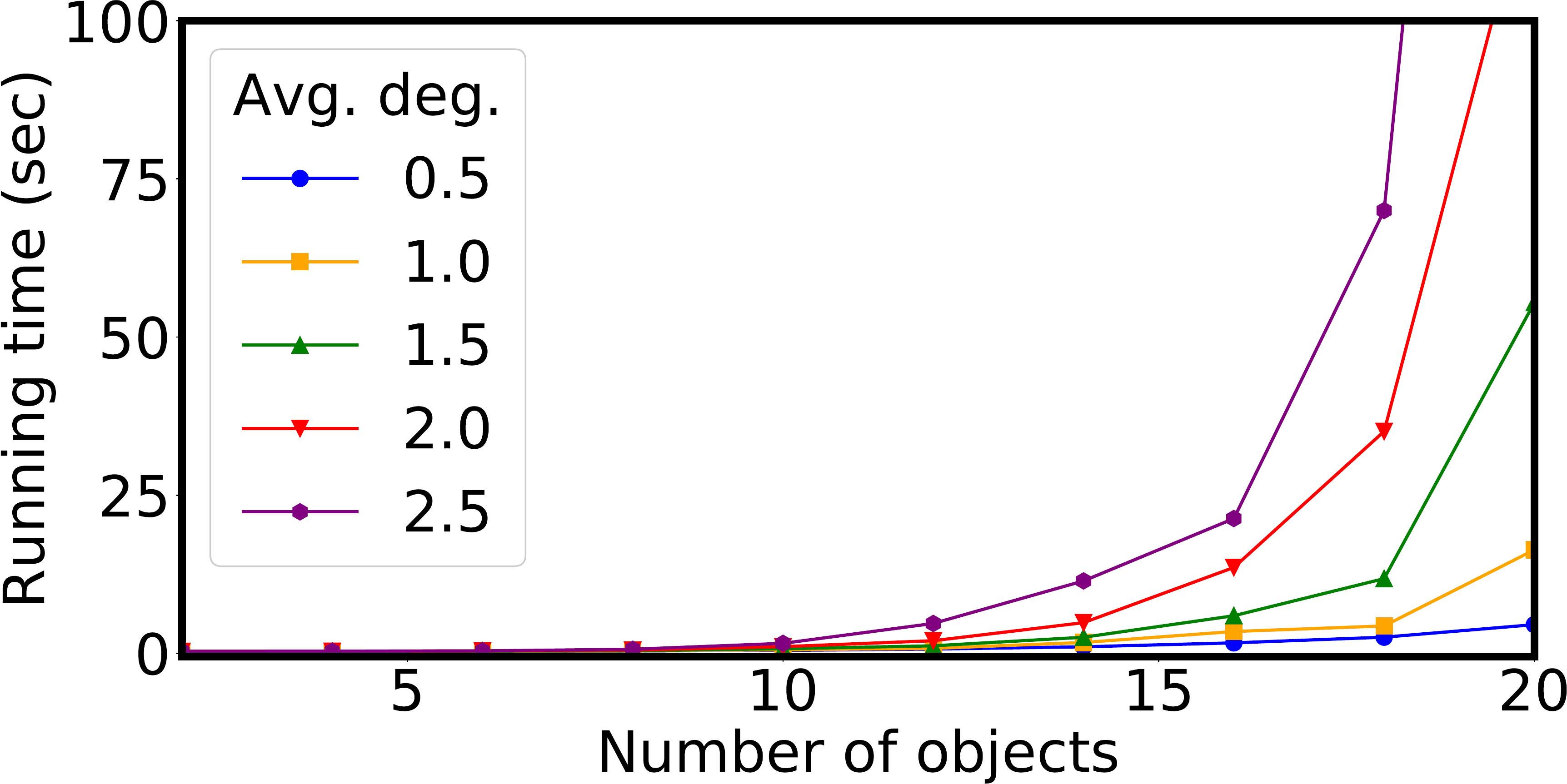}
	\caption{The total running time for {\sc ToroFVSSingle}().}
	\label{fig:lwo-total}
\end{figure}

The running time for the entire {\sc ToroFVSSingle}() is provided 
in Fig.~\ref{fig:lwo-total}. Observe that FVS computation takes 
almost no time in comparison to the distance minimization step. 
As expected, higher average degrees in $G_{dep}$ make the computation 
harder.

\begin{figure}[htp]
		\centering
		\includegraphics[keepaspectratio, width = .77\columnwidth]{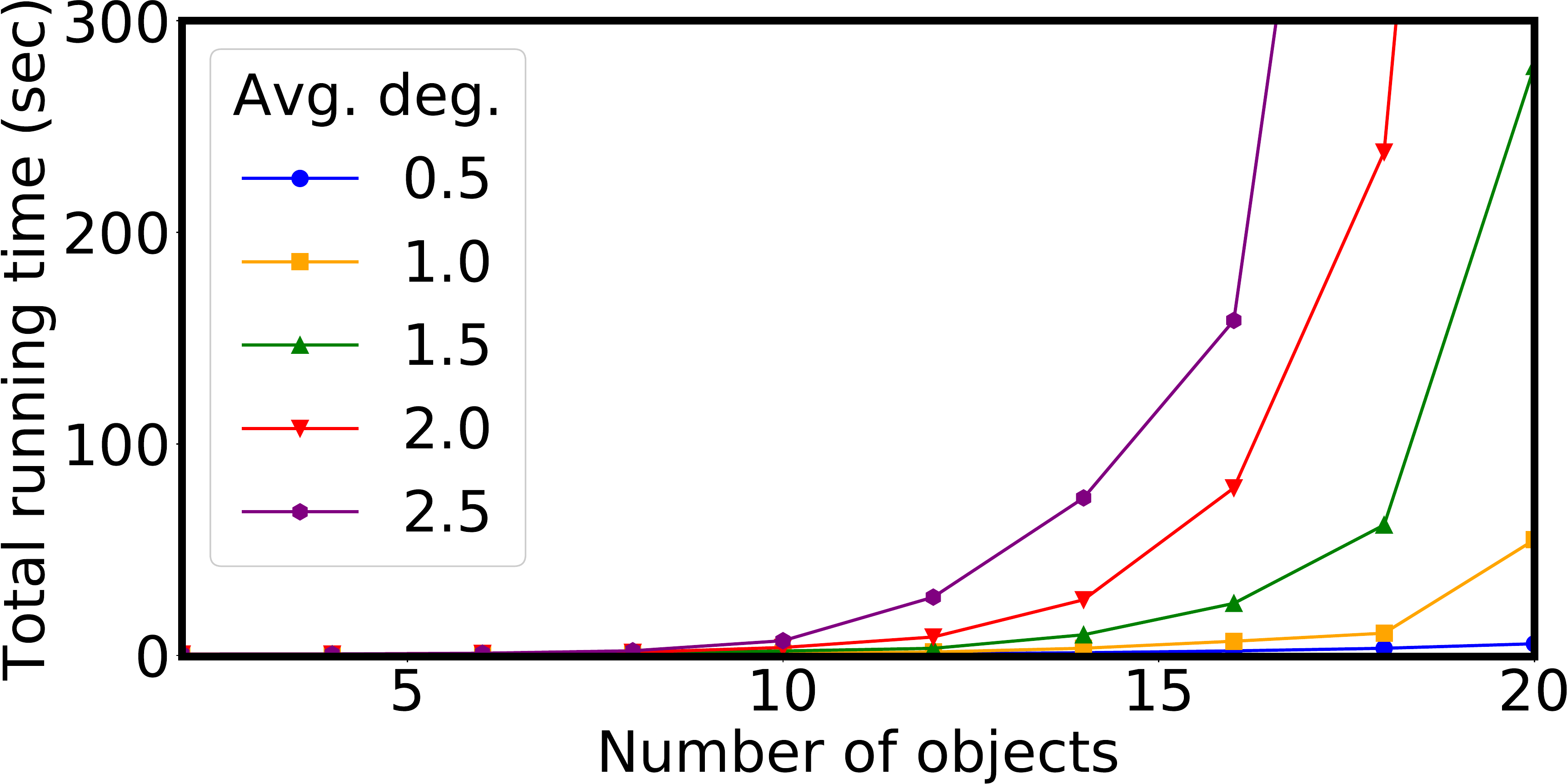}
		\caption{The running time to produce a global optimal solution for \rwo.}
		\label{fig:rwo-global-total}
	\end{figure}

Running {\sc ToroFVSSingle}() together with FVS enumeration, a global
optimal solution is computed for \rwo under the assumption that the
grasp/release costs dominate. Only solutions with an optimal FVS are
considered. The computation time is provided in
Fig.~\ref{fig:rwo-global-total}. The result shows that it gets costly
to compute the global optimal solution as the number of objects go
beyond $15$ for dense setups. It is empirically observed that for the
same problem instance and different optimal FVSs, the minimum distance
computed by {\sc MinDist}() in Alg.~\ref{algo:lwo} has less
than $5\%$ variance.  This suggests that running {\sc ToroFVSSingle}()
just once should yield a solution that is very close to being the
global optimum.

\section{Physical Experiments}\label{sec:hardware-exp}

\begin{figure}[htp]
    \centering
    \includegraphics[keepaspectratio, width = 0.9\linewidth]{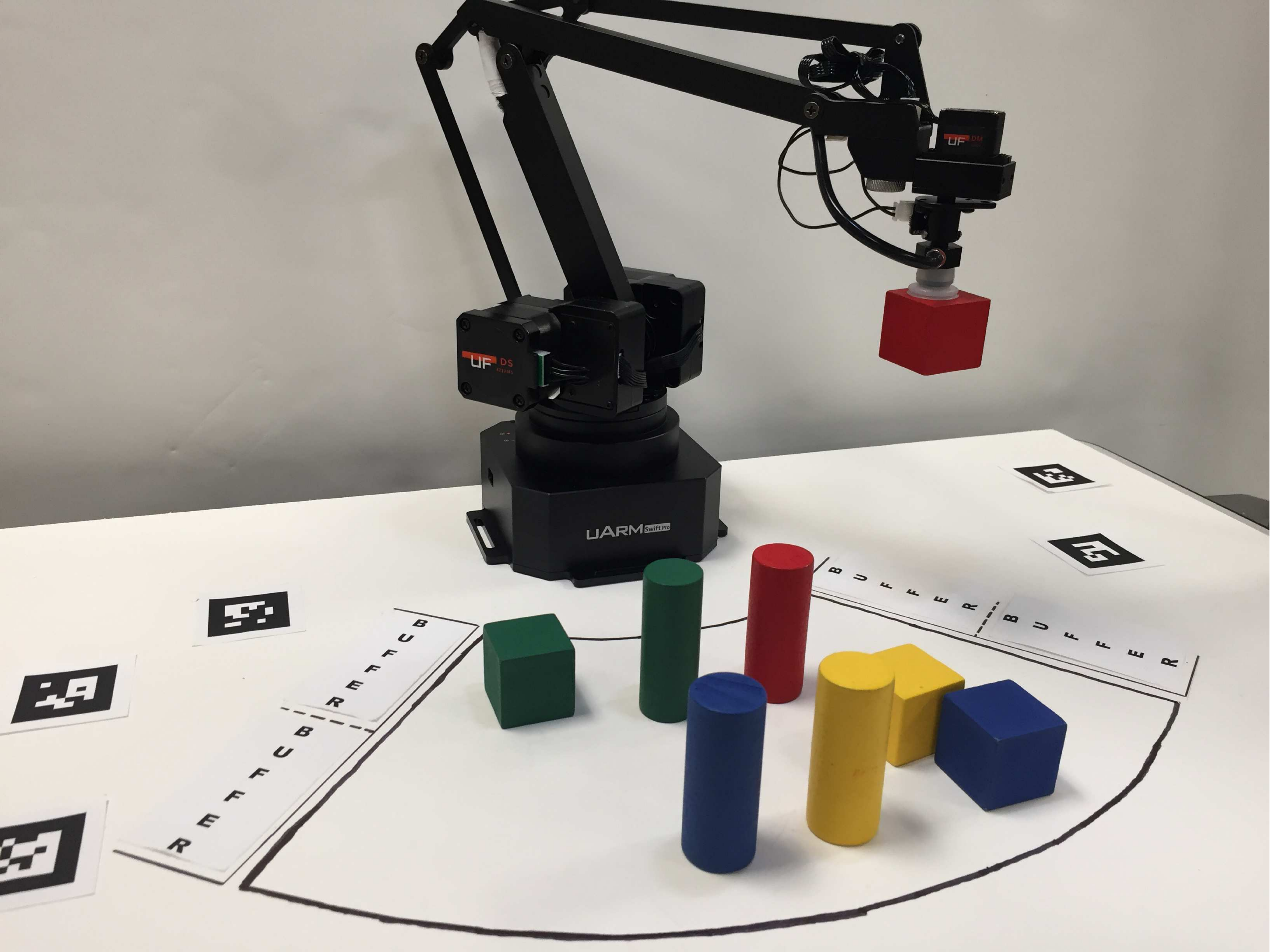}
    \caption{A snapshot of the physical system carrying out a solution
      generated by the algorithms presented in this paper for a
      tabletop rearrangement scenario.}
    \label{fig:hardware-total}
\end{figure}

This section demonstrates an implementation of the algorithms
introduced in this paper on a hardware platform
(Fig.~\ref{fig:hardware-total}).  The physical experiments were
conducted on three different tabletop scenarios. For each scenario,
multiple problem instances were generated to compare the efficiency of
the solutions produced by the proposed algorithms relative to
alternatives, such as random and greedy solutions.

In order to apply object rearrangement algorithms in the physical
world, it is first necessary to identify the problem parameters
corresponding to the formulation of the \toro problem.  Specifically,
the algorithms are expecting as input a start and goal arrangement on
a tabletop. The goal arrangement is predefined, while the objects are
in an initial, arbitrary state. The physical platform first detects
the starting arrangement of the objects before executing a solution
for rearranging the objects into the desired goal arrangement.

\subsection{Hardware Setup}
\noindent The experimental setup is comprised of five components:

\newcounter{component}
\stepcounter{component}

\begin{figure}[!ht]
    \centering
    \includegraphics[keepaspectratio, width = 0.99\columnwidth]{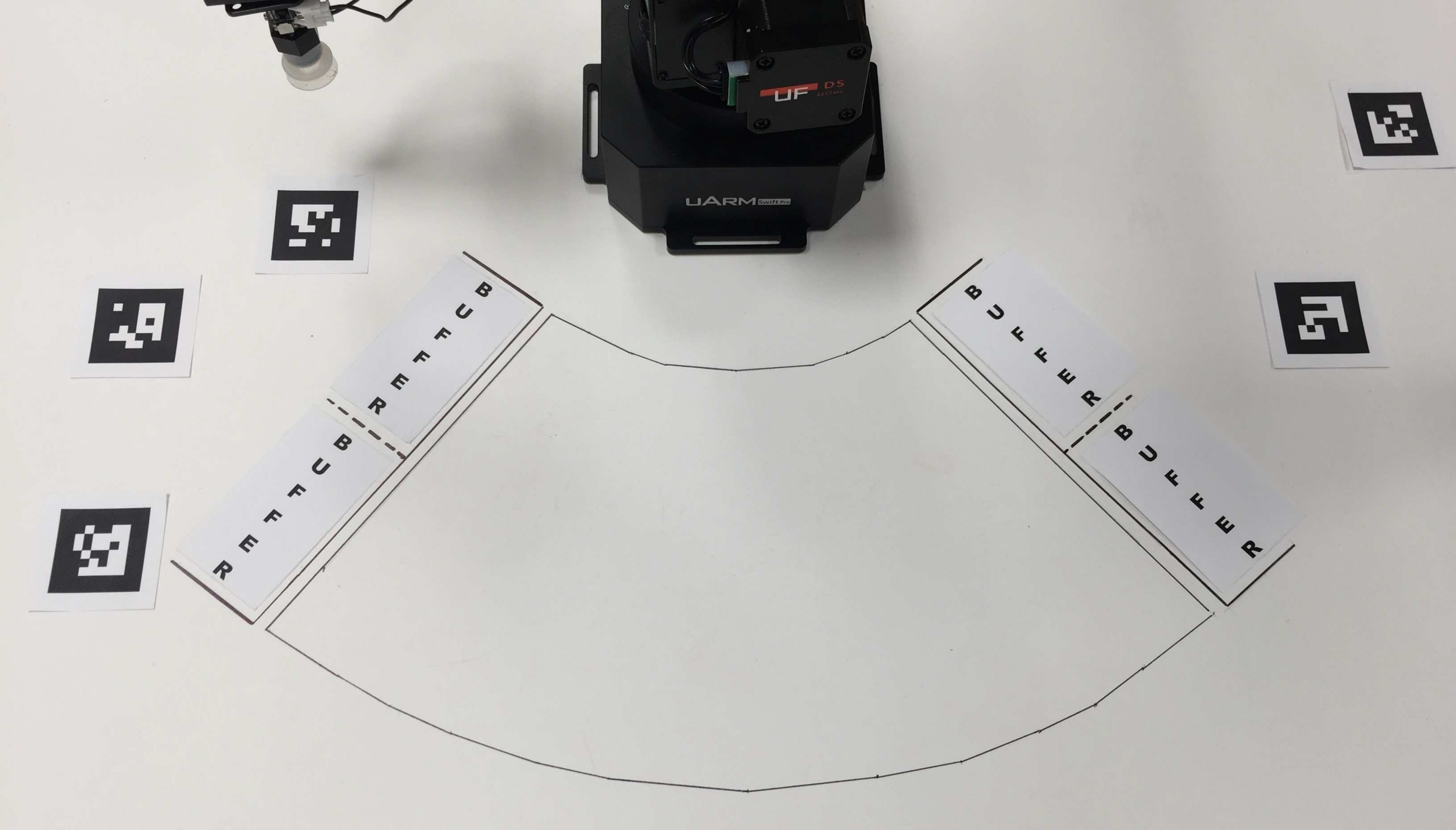}
    \caption{A tabletop environment where the uArm Swift Pro's
      workspace is the area between $[45^{\circ}, 135^{\circ}]$ at a
      distance ranging between 140mm and 280mm from the base.}
    \label{fig:experiment-tabletop}
\end{figure}

\noindent\textbf{\arabic{component}.~Tabletop} The tabletop is a
planar surface where the desktop manipulator and objects rest.
Clearly defined contours are drawn denoting the projections of the
manipulator's reachable area (i.e., workspace), including predefined
buffers within this reachable area, as shown in
Fig~\ref{fig:experiment-tabletop}.

\begin{figure}[htp]
    \begin{tabularx}{\linewidth}{Zc}
    \includegraphics[keepaspectratio, height=1.23in]{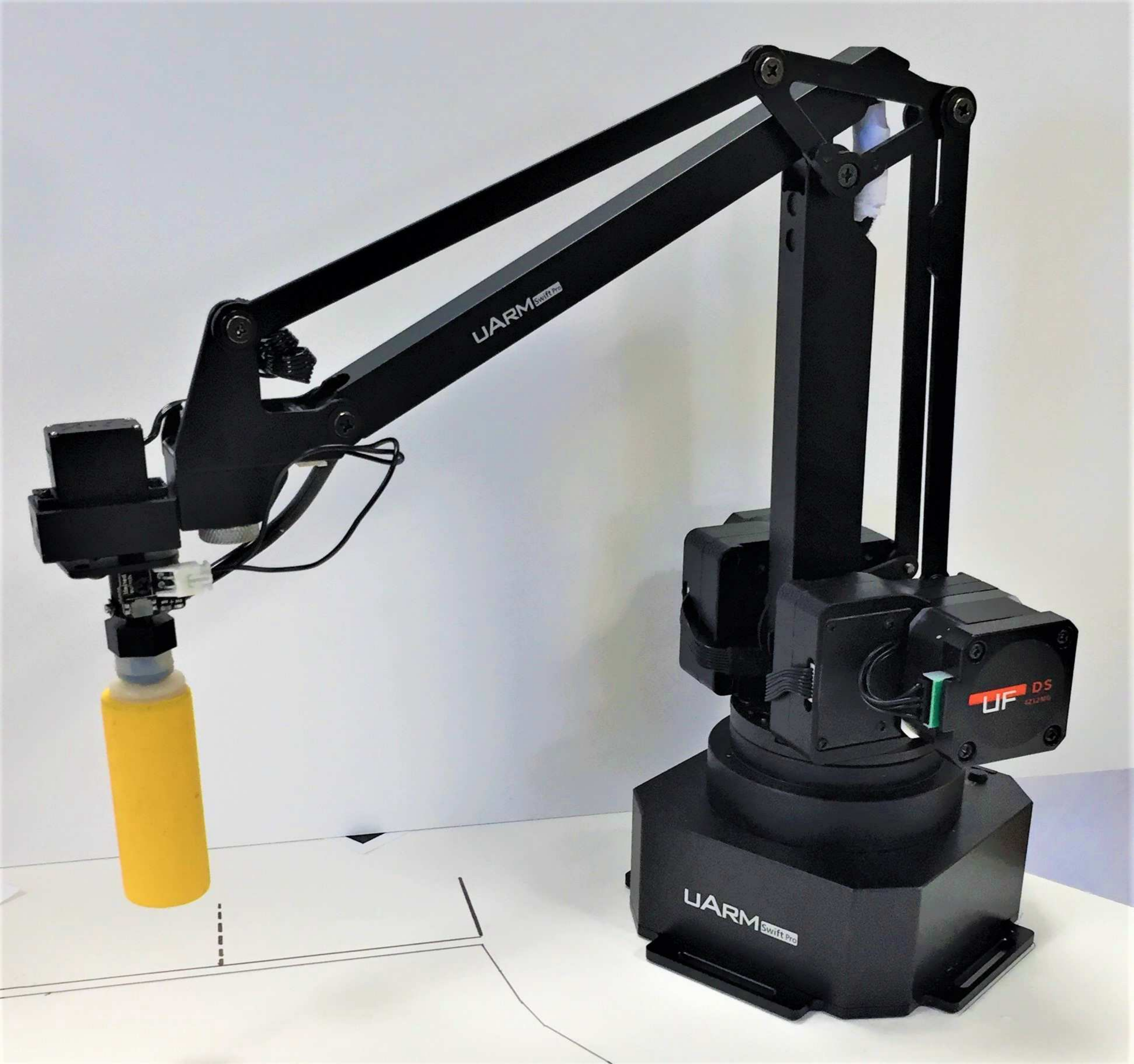} & 
    \includegraphics[keepaspectratio, height=1.23in]{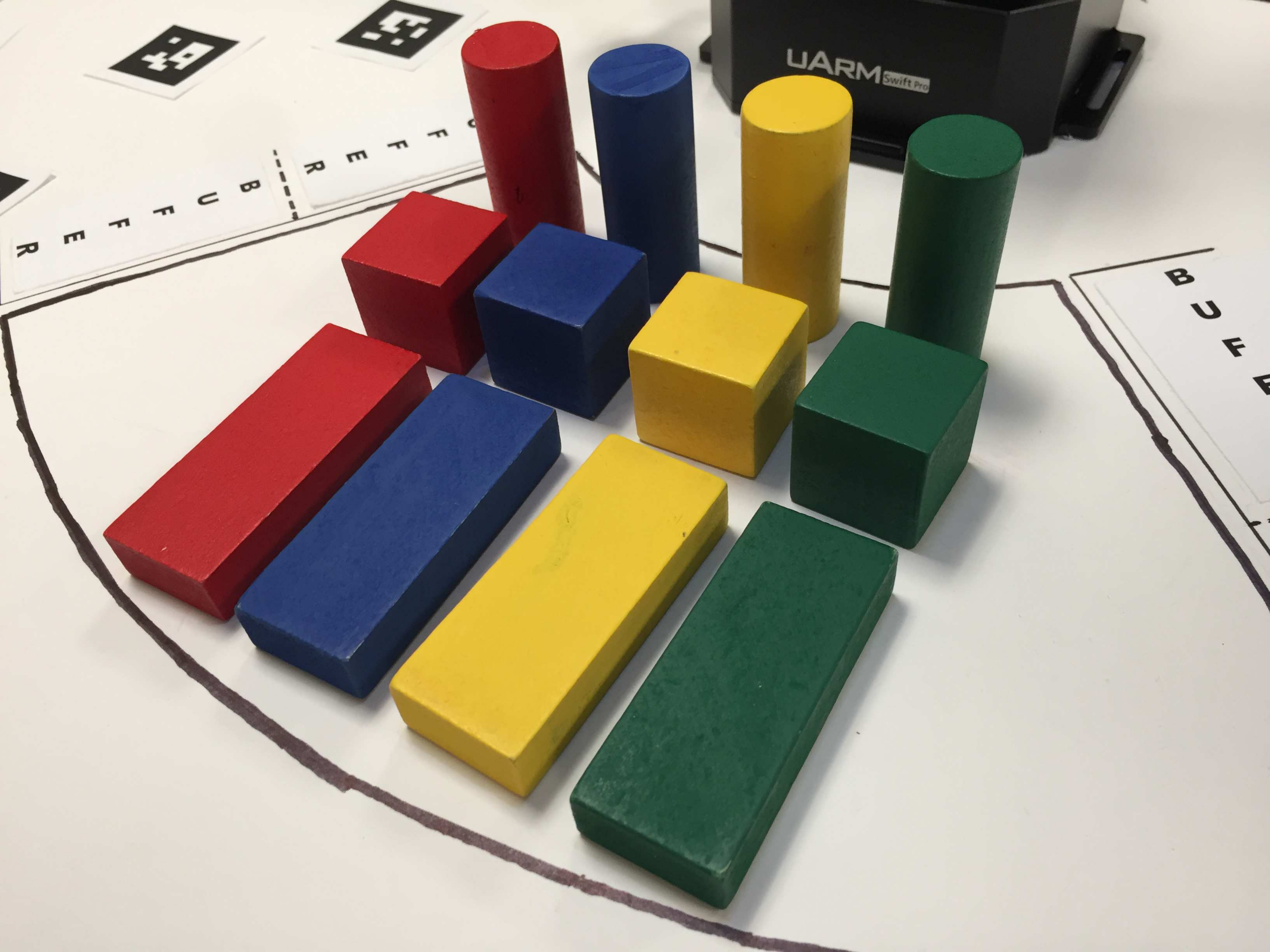} \\
    (a) & (b) \\
    \end{tabularx}
    \caption{(a) The UFACTORY uArm Swift Pro grasping one of the labeled (colored) cylindrical objects.
    (b) An arrangement of 12 objects of various color and geometry.}
    \label{fig:manipulator}
\end{figure}

\stepcounter{component}
\noindent\textbf{\arabic{component}.~Manipulator:} The UFACTORY uArm
Swift Pro \footnote{\url{http://www.ufactory.cc/}. The authors would
  like to thank uFactory for supplying the uArm Swift Pro robot that
  was used in the hardware-based evaluation.}
(Fig.~\ref{fig:manipulator}a) is an inexpensive but versatile desktop
manipulator capable of performing repeatable actions with a precision
of 0.2mm. The uArm Swift Pro's versatility is largely due to the
variety of end-effectors that can be equipped.  The suction cup
end-effector is utilized to achieve overhand grasps in the forthcoming
challenges. Although the uArm Swift Pro is not built for industrial
applications, it has enough precision to perform the experimental
tasks described in this section.

\stepcounter{component}
\noindent\textbf{\arabic{component}.~Camera:} The Logitech\copyright
webcam C920
\footnote{\url{https://www.logitech.com/en-us/product/hd-pro-webcam-c920}}
is a consumer grade webcam capable of Full HD video recording ($1080$p
- $1920 \times 1080$ pixels) that is used for object pose detection in
the experiments. The camera is mounted above the tabletop, such that
it's field-of-view~(Fig.~\ref{fig:experiment-tabletop}) captures the
entire workspace.  The pre-calibration eliminates lens distortion and
also renders brighter, saturated images, which makes pose detection
easier.
 
\stepcounter{component}
\noindent\textbf{\arabic{component}. Marker Detection} ``Chilitags'' is
a cross-platform software library for the detection and identification
of two-dimensional fiducial markers \citep{BonSev+13}.  Physical
markers placed in the view of the imaging system are used as a point
of reference/measure for surrounding objects.  In the case of
Chilitags, that object is a physical marker added to the environment.
As seen in Fig.~\ref{fig:experiment-tabletop}, the experiments employ
five tags, each placed at a known pose on the tabletop. This knowledge
facilitates the computation of the 2D transformation between the
manipulator's frame of reference and the camera's frame of reference.

\stepcounter{component}
\noindent\textbf{\arabic{component}.~Objects} There are several
objects that the manipulator interacts with during the experiments
(Fig.~\ref{fig:manipulator}b). These objects are identifiable
according to their shape and color \vspace{-.1in}$$\{\text{cube,
  cylinder, orthotope}\} \times \{\text{red, blue, green,
  yellow}\} \vspace{-.1in}$$ resulting in twelve uniquely identifiable
objects. The center of each object on the tabletop and its orientation
define its configuration.

\subsection{Object Pose Detection}

The predetermined goal configuration of the objects is defined
relative to the manipulator's frame of reference.  The initial
configuration is unknown and is determined online via the overhead
C920 camera.  Once the objects are detected, their observed
configuration undergoes a transformation to the manipulator's frame of
reference.

The pose estimation method appears in Alg.~\ref{algo:pose-estimation}.
In line~\ref{algo:pose-estimation-camera}, an image is taken by the
camera. Line~\ref{algo:pose-estimation-blur} utilizes \textsl{Gaussian
  blur} (a.k.a.  Gaussian smoothing) to reduce image noise.  For each
of the predefined colors (i.e., red, blue, yellow, green), the area(s)
of the image matching the current color are extracted and further
smoothed by \textsl{morphological transformations}, including
\textsl{erosion} and \textsl{dilation}.  The contours of these areas,
which describe the top sides of objects, are then calculated
(Line~\ref{algo:pose-estimation-contours}).  Each contour is examined
to determine whether or not it corresponds to one of the objects in
the scene (line~\ref{algo:pose-estimation-shape}).  For each of the
contours corresponding to an object in the scene, several operations
need to occur to determine pose of the objects.  The 2D point
component of the pose as it appears in the camera is then determined
by computing the center of the contour's minimum enclosing circle
(line~\ref{algo:pose-estimation-location}), while the orientation of
the object in the camera's frame is extracted via \textsl{principle
  component analysis} over the minimum area rectangle containing the
contour (line~\ref{algo:pose-estimation-orientation}).
Line~\ref{algo:pose-estimation-transformation} updates the 2D point
component of the pose in the camera's frame to accurately reflect
position of the object relative to the manipulator, taking into
account the current shape geometry.  The 2D perspective transformation
between camera frame and robot frame is pre-computed using the marker
detection software.  The poses of the tags in the robot's frame are
fixed, but the pose of the tags in the camera's frame are
automatically detected at runtime.

\begin{algorithm}
    \small
	\DontPrintSemicolon
    $objects \gets \{\}$\;
    $img \gets \text{\sc CameraCapture}()$\;\label{algo:pose-estimation-camera}
    $img \gets \text{\sc GaussianBlur}(img)$\;\label{algo:pose-estimation-blur}
    \For{$color \in \{red, blue, yellow, green\}$}{\label{algo:pose-estimation-color}
        $contours \gets \text{\sc FindContours}(img, color)$\;\label{algo:pose-estimation-contours}
        \For{$contour \in contours$}{
            $shape \gets \text{\sc DetectShape}(contour)$\;\label{algo:pose-estimation-shape}
            \If{$shape \in \{cube, cylinder, orthotope\}$}{
                $position \gets \text{\sc Center}(\text{\sc MinEnclosingCircle}(contour))$\;\label{algo:pose-estimation-location}
                $orientation \gets \text{\sc PCA}(\text{\sc MinAreaRect}(contour))$\;\label{algo:pose-estimation-orientation}
                $pose \gets \text{\sc Update}(position, orientation, shape)$\;\label{algo:pose-estimation-transformation}
                \If{\text{\sc InWorkspace}($pose$)}{
                    $objects \gets objects \cup \{(shape, color, pose)\}$
                }
            }
        }
    }
	\Return{$objects$}\;
	\caption{{\sc ObjectPoseEstimation}}
	\label{algo:pose-estimation}
\end{algorithm}

\subsection{Experimental Validation}\label{ssect:ev}

This section presents three tabletop object rearrangement scenarios
that can be performed via overhand pick-and-place actions.  For each
scenario, specific problem instances have been provided that are
solvable.  The specific algorithm is determined at run-time, and
corresponds to whether there is overlap between the start and goal
object configurations.  If a subset of the start and goal
configurations overlap, the problem may require the use of external
buffer(s).  The number of extra pick-and-place actions and the
corresponding number of buffers necessary to carry out the task, can
be determined via the dependency graph.

All of the generated solutions by the proposed methods perform an
optimal number of pick-and-place actions.  A solution is optimal with
respect to the travel distance of the end-effector when the external
buffers are not utilized. Problem instances that require the use of an
external buffer(s) remain near optimal with respect to the travel
distance of the end-effector.

For each problem instance, a feasible solution is also generated by
either a random algorithm (for \rno) or a greedy algorithm (for
\toro\footnote{The solution produced by the \lwoalgs
algorithm uses the solution returned by the Gurobi ILP solver 
after 10 sec of compute time. Empirically, 
any further computation only minimizes the transition time between poses 
at the expense of increased computation. 
Note that this does not affect the number of grasps which remain optimal.}).
The execution time for the different solutions are then
measured and compared.


\begin{figure}[!h]
    \begin{tabularx}{\linewidth}{@{}>{\centering\arraybackslash}X@{}>{\centering\arraybackslash}X@{}}
       \includegraphics[width = .98\linewidth]{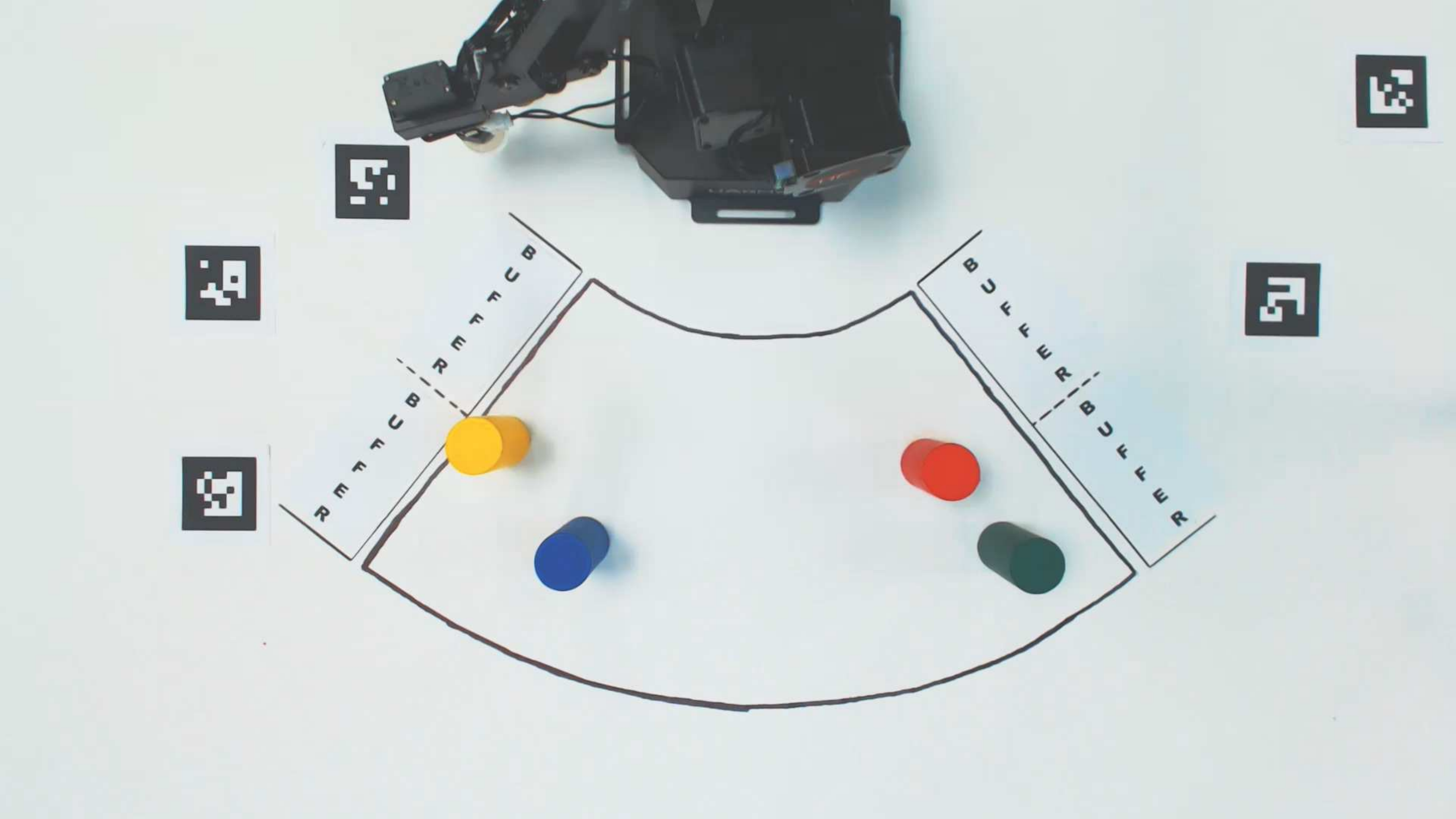} & 
       \includegraphics[width = .98\linewidth]{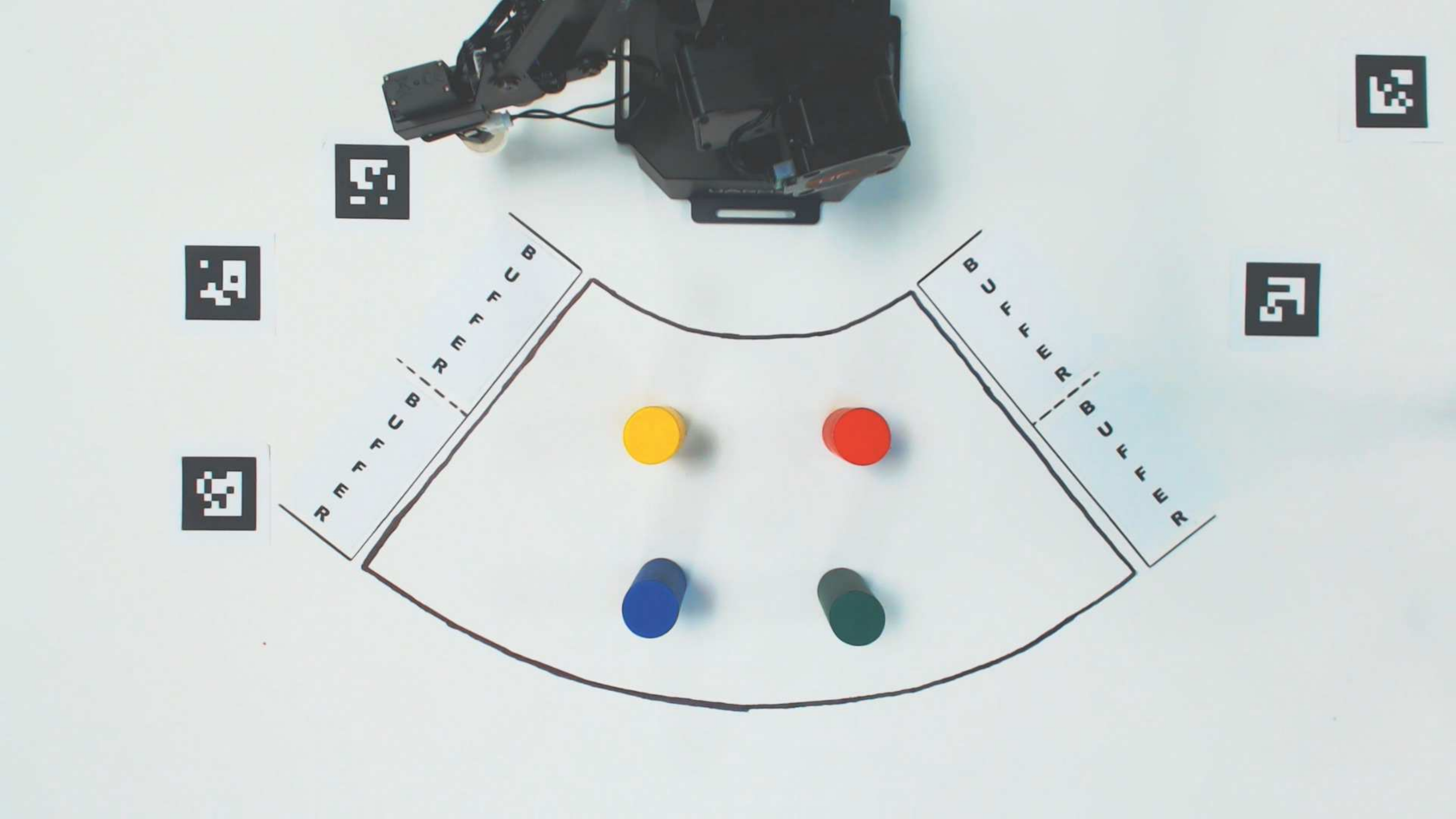} \\
       (a) Start  arrangement & (b) Goal arrangement 
    \end{tabularx}
    \caption{Problem instance of Scenario~1.}
    \label{fig:experiment-1}
\end{figure}

\paragraph{Scenario 1: Cylinders without Overlap} This task requires
the manipulator arm to transport four uniquely identifiable cylinders
from a random start configuration to a fixed goal configuration.
Figures~\ref{fig:experiment-1}a and \ref{fig:experiment-1}b illustrate
one such problem instance where the start and goal configurations do
not overlap, indicating that this is a \rno instance.  Since \rno does
not necessitate the use of any external buffers, the manipulator
perform only four pick-and-place operations, corresponding to the
number of cylinders in the scene.  Appendix~\ref{app:exp-sol} provides
the \textsl{distance optimal sequence} for pick-and-place operations
of this instance.

To illustrate the efficiency of solutions generated by \rnos, 10
different problem instances are generated. They are solved by both
\rnos~and random grasping sequences. Note that any permutation of
objects is a feasible solution.  The result is presented in
Table~\ref{table:hardware-no}.  The first column in this table
specifies different problem instances.  The \textsl{raw} columns
measure the total execution time, which contains grasps, releases, and
transportation.  Since the time for grasps and releases is the same
for all the feasible solutions for a \rno~instance, the amount of time
($45.23$ sec) to perform four in-place pick-and-place actions is then
deducted from total execution time.  These results appear in the
\textsl{tran} columns.  Moreover, the difference between the execution
time of the algorithms appear in the \textsl{difference} column.

\begin{table}
    \small
    \begin{tabularx}{\linewidth}{@{}|Z|c|c|c|c|c|@{}}
    \hline
    \multirow{3}{*}{Instance} & \multicolumn{5}{c|}{Execution Time (sec)} \\ \cline{2-6}
    & \multicolumn{2}{c|}{\rnos} & \multicolumn{2}{c|}{Random Solution} & \multirow{2}{*}{Difference} \\ \cline{2-5}
        & Raw & Tran & Raw & Tran & \\ \hline
    1	& 65.04	& 19.81	& 68.50 & 23.27 & 3.46 \\ \hline
    2	& 70.44	& 25.21	& 75.38 & 30.15 & 4.94 \\ \hline
    3	& 67.11	& 21.88	& 69.85 & 24.62 & 2.74 \\ \hline
    4	& 69.34	& 24.11	& 73.09 & 27.86 & 3.75 \\ \hline
    5	& 70.67	& 25.44	& 73.04 & 27.81 & 2.37 \\ \hline
    6	& 71.13	& 25.90	& 73.37 & 28.14 & 2.24 \\ \hline
    7	& 73.40	& 28.17	& 73.66 & 28.43 & 0.26 \\ \hline
    8	& 67.95	& 22.72	& 67.97 & 22.74 & 0.02 \\ \hline
    9	& 66.62	& 21.39	& 68.21 & 22.98 & 1.59 \\ \hline
    10	& 66.88	& 21.65	& 68.22 & 22.99 & 1.34 \\ \hline
    Average	& 68.86	& 23.63	& 71.13 & 25.90 & 2.27 \\ \hline
    \end{tabularx}
    \caption{Experimental results showing a comparison of the the execution
    time between \rnos and a random feasible solution for ten problem
    instances.}
    \label{table:hardware-no}
\end{table}

From the empirical data provided in Table~\ref{table:hardware-no}, 
random solution strategies for this problem setup incur on average a 
$9.6\%$ increase in transportation time compared to the solution 
generated by \rnos.

\begin{remark}
    The difference of execution time between a random solution and 
    the optimal solution is generally proportional, but not linearly 
    dependent to the transition cost defined in this paper, 
    which is based on the Euclidean distance between poses in 2D. 
    This is due to the manipulator model. 
    In practice, the path that the end-effector travels does not necessarily need 
    to be along piece-wise linear shortest paths (straight lines). 
    Additionally, the second order term (acceleration) varies, contributing to 
    a non-static velocity, which is not captured by the tabletop model described herein.
\end{remark}

\begin{table*}[t]
    \centering
    \small
    \begin{tabular}{|c|c|c|c|c|}
    \hline
    \multirow{2}{*}{Scenario} & \multicolumn{2}{c|}{Execution Time (sec)} & \multicolumn{2}{c|}{Num.~of Actions} \\ \cline{2-5}
    & \lwoalgs & Greedy Alg. & \lwoalgs & Greedy Alg. \\ \hline 
    2 & 165.57 & 190.02 & 9 & 10 \\ \hline
    3 & 299.95 & 405.16 & 16 & 20 \\ \hline
    \end{tabular}
    \caption{Experimental results showing a comparison of the the execution
    time between \lwoalgs and a greedy solution for two scenarios.}
    \label{table:hardware-wo}
\end{table*}


\begin{figure}[!h]
    \begin{tabularx}{\linewidth}{@{}>{\centering\arraybackslash}X@{}>{\centering\arraybackslash}X@{}}
       \includegraphics[width = 0.98 \linewidth]{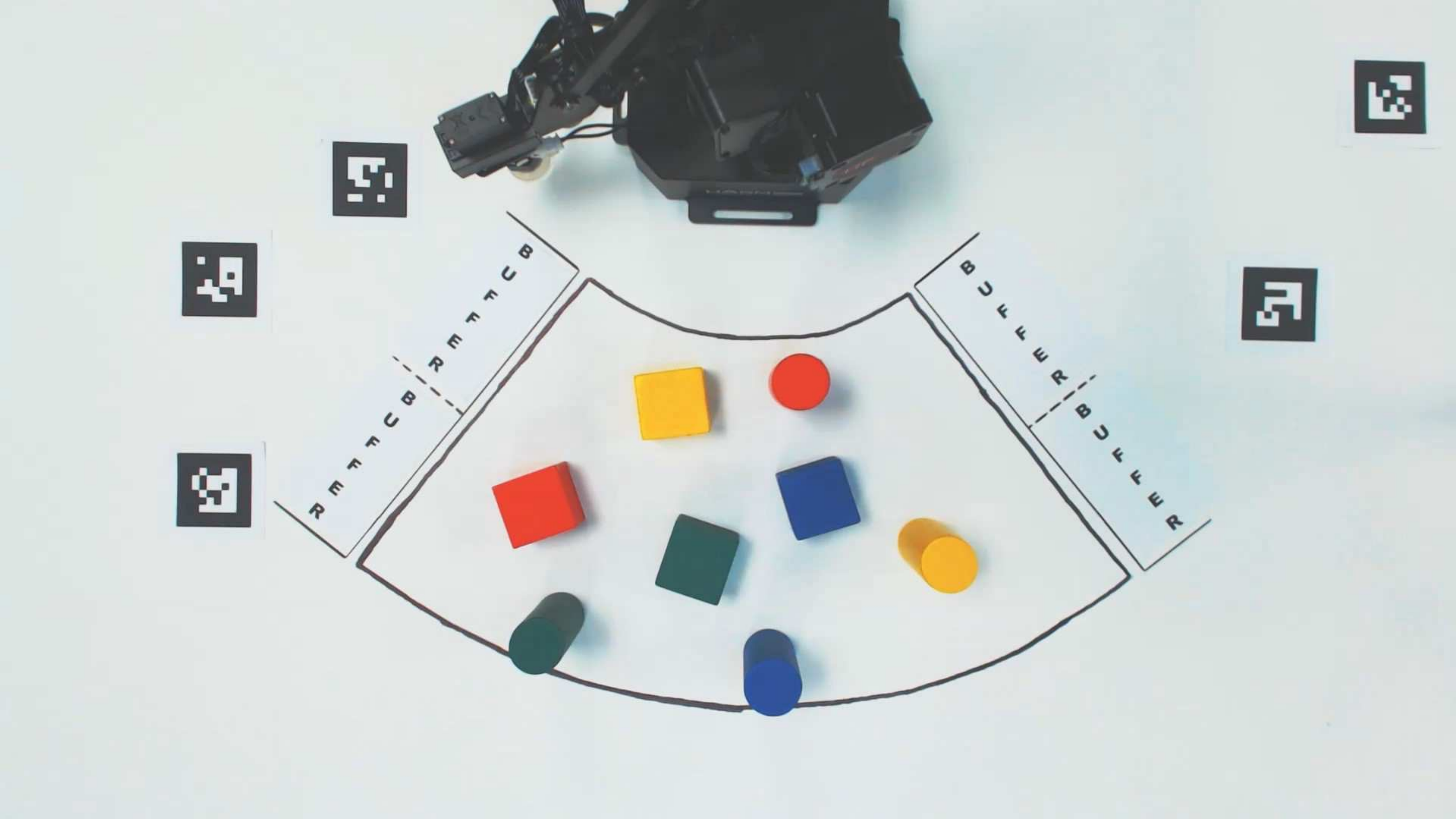} & 
       \includegraphics[width = 0.98 \linewidth]{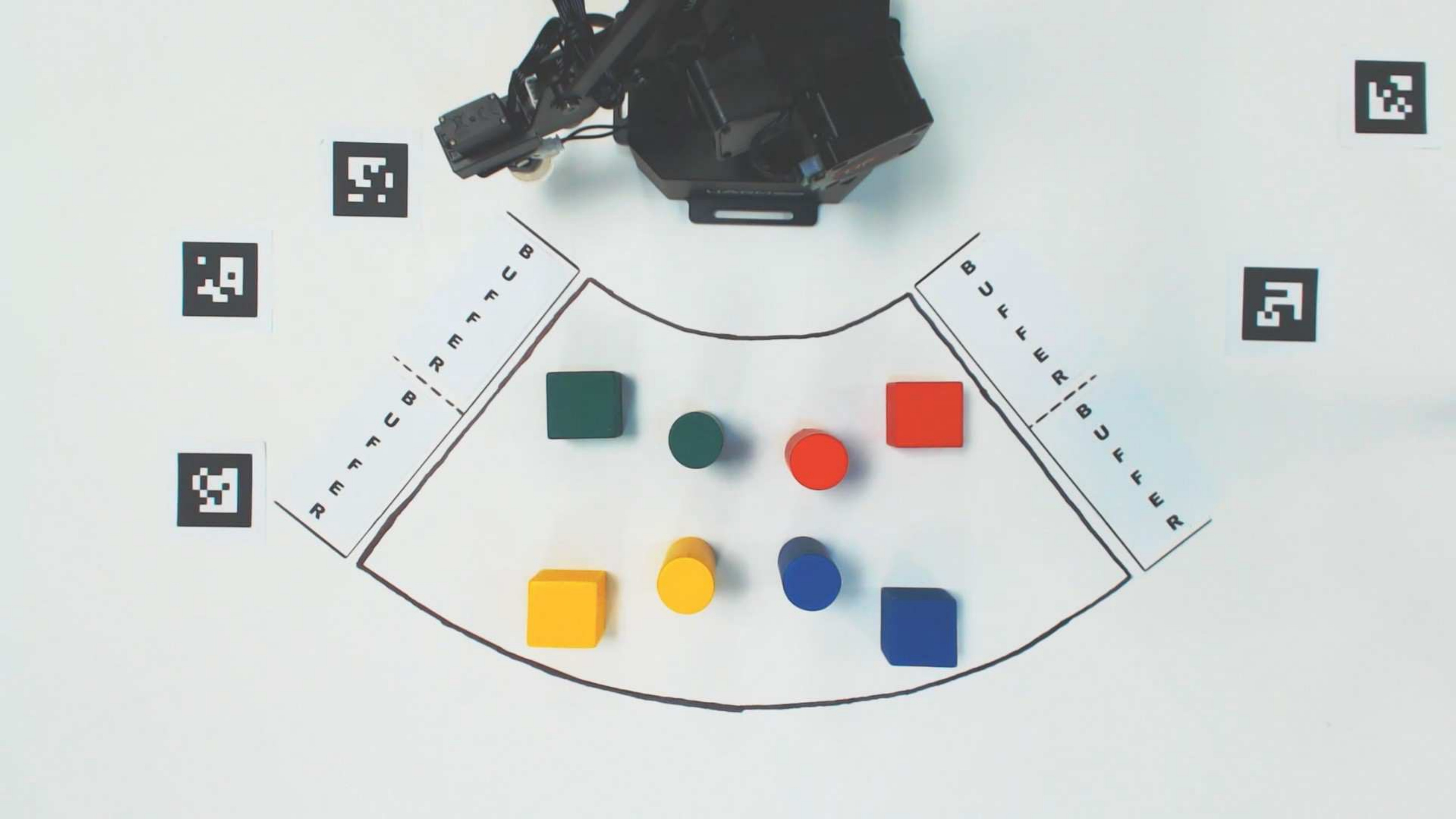} \\
       (a) Start  arrangement & (b) Goal arrangement 
    \end{tabularx}
    \caption{Problem instance of Scenario~2.}
    \label{fig:experiment-2}
\end{figure}

\paragraph{Scenario 2: Cylinders and Cubes} In this scenario, the
arm is tasked with rearranging eight objects (four cylinders and four
cubes) that are initially scattered throughout the workspace, while
the desired goal configuration is pre-determined.
Figures~\ref{fig:experiment-2}a~and~\ref{fig:experiment-2}b show one
such instance.  Due to overlap between the start and goal
configurations, this is a \toro instance and thus uses the \lwoalgs
algorithm, making use of the external buffers available to the
manipulator.  In the solution of \lwoalgs on this particular instance,
the manipulator uses one of the available buffer locations to perform
the rearrangement.  The movement of an object to an external
buffer results in a total of nine pick-and-place actions.

\lwoalgs is compared to a greedy method, which solves the problems
sequentially by first removing the dependencies for one object and
then it moves it to its goal.  As shown in Table
\ref{table:hardware-wo}, the greedy algorithm returns $1$ extra grasps
and takes $24.45$ secs of additional execution time, compared to
\lwoalgs.


\begin{figure}[!h]
    \begin{tabularx}{\linewidth}{@{}>{\centering\arraybackslash}X@{}>{\centering\arraybackslash}X@{}}
       \includegraphics[width = 0.98 \linewidth]{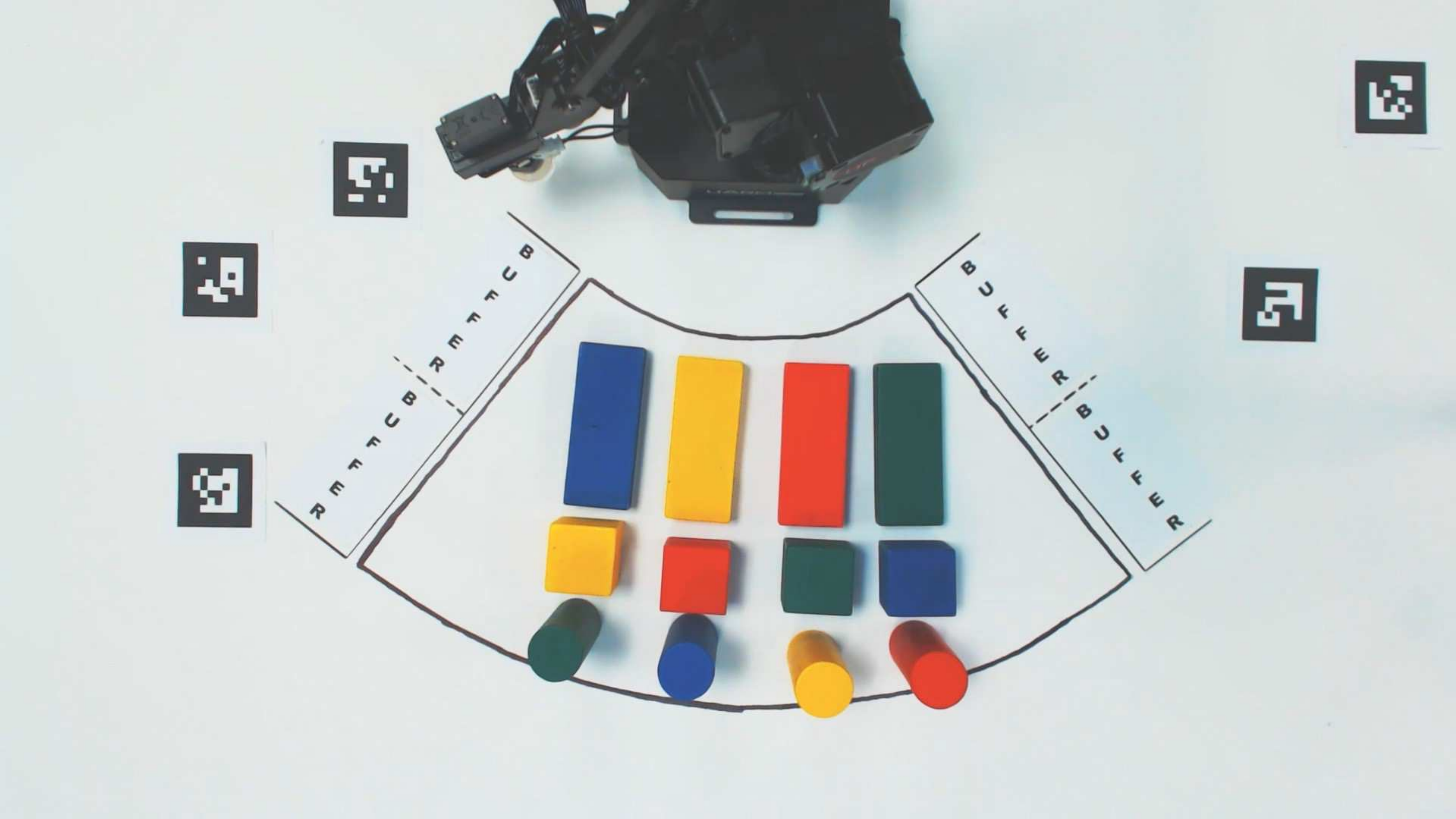} & 
       \includegraphics[width = 0.98 \linewidth]{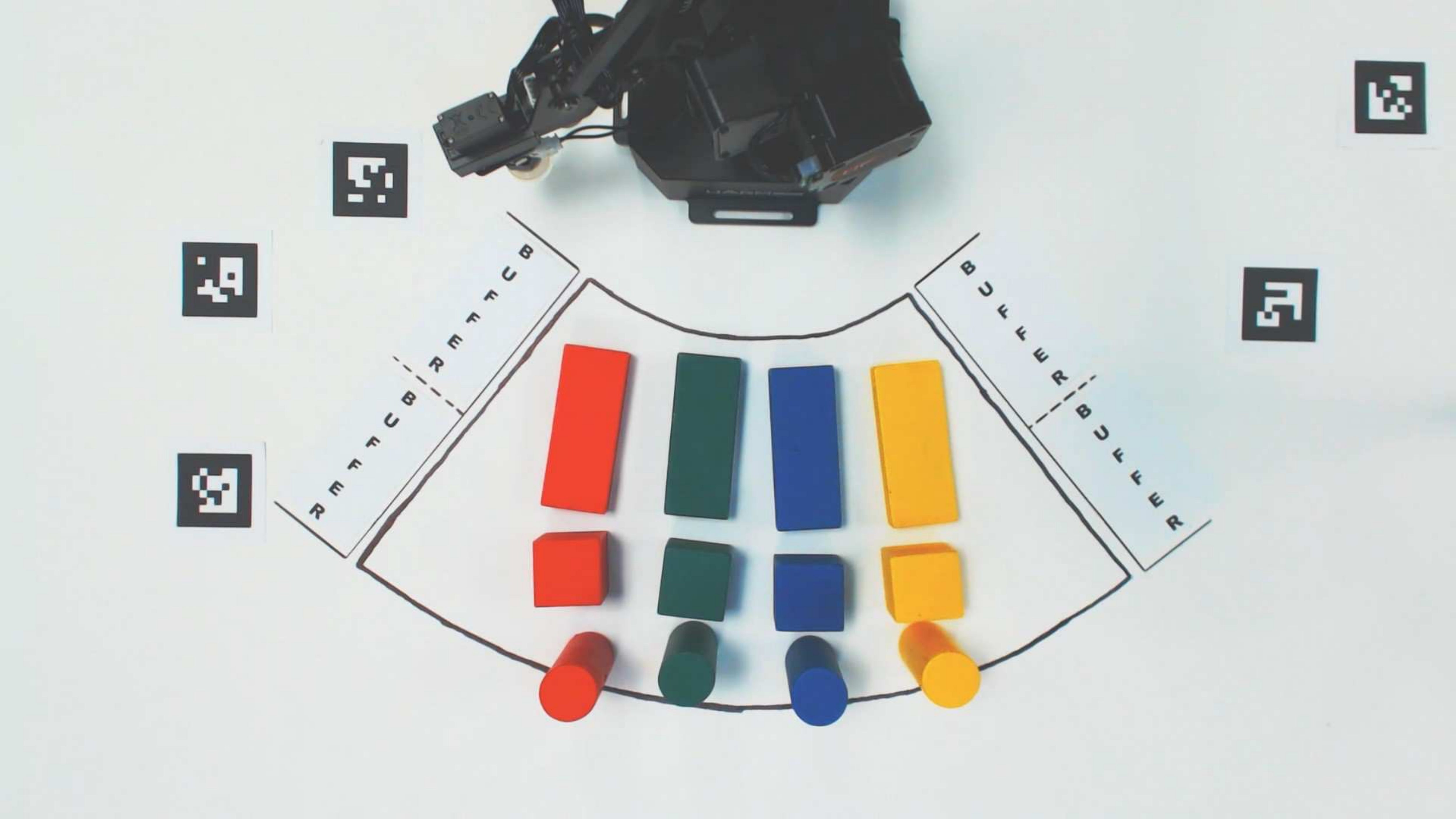} \\
       (a) Start arrangement & (b) Goal arrangement 
    \end{tabularx}
    \caption{Problem instance for Scenario~3 containing twelve objects.}
    \label{fig:experiment-3}
\end{figure}

\paragraph{Scenario 3: Cylinders, Cubes, Orthotopes} This scenario
involves rearranging twelve objects (four of each type of cylinders,
cubes and orthotopes) within the manipulator's workspace.  Objects of
identical geometry are aligned in front of the manipulator, ordered by
increasing height (i.e., orthotopes, cubes, cylinders).  Thus, start
and goal configurations differ by permutations of color amongst
objects of the same shape.  The instance shown in
Figures~\ref{fig:experiment-3}a~and~\ref{fig:experiment-3}b utilizes
three of the available buffer locations as it performs sixteen
pick-and-place actions necessary to solve the task.

As the number of objects increases, the difference between the
solution quality of \lwoalgs and the greedy algorithm becomes larger.
The solution of greedy algorithm has $4$ extra grasps and $105.21$
secs more execution time compared to the sub-optimal solution
generated by \lwoalgs.


\section{Conclusion} 
\label{sec:conclusion}

This paper studies the combinatorial structure inherent in tabletop
object rearrangement problems. For \rno and \uno, it is shown that
Euclidean-\tsp can be reduced to them, establishing their
NP-hardness. More importantly, \rno and \uno can be reduced to \tsp
with little overhead, thus establishing that they have similar
computational complexity and lead to an efficient solution
scheme. Similarly, an equivalence was established between dependence
breaking of \rwo and \fvs, which is APX-hard. The equivalence enables
subsequent ILP-based methods for effectively and optimally solving
\rwo instances containing tens of objects with overlapping starts and
goals.
	
The methods and algorithms in this paper serve as an initial
foundation for solving complex rearrangement tasks on tabletops.  Many
interesting problems remain open in this area; two are highlighted
here.  The current paper assumes the availability of {\em external}
buffers, which are separated from the workspace occupied by the
objects, demanding additional movement from the end-effector.  In
practice, it can be beneficial to dynamically locate buffers that are
close by, which may be tackled through effective sampling methods.
Furthermore, the scenarios addressed only static settings whereas many
industrial settings require solving a more dynamics problem in which
the objects to be rearranged do not remain still with respect to the
robot base.

\appendix
\section{Proof for Cost-optimal \uno}\label{proof-couno}

\begin{proof}[Proof of Theorem IV.2]
Again, reduce from Euclidean-\tsp. The same \tsp instance from the proof of
Theorem~IV.1 is used. The conversion to \rno and the process to
obtain a \uno instance are also similar, with the exception being that edges
$s_ig_i$ are not required to be used in a solution; this makes the labeled case
become unlabeled. 

The argument is that the cost-optimal solution of the \uno instance also yields
an optimal solution to the original Euclidean-\tsp tour. This is accomplished
by showing that an optimal solution to the \rno instance has essentially the
same cost as the \uno instance. To see that this is the case, assume that an
optimal solution (tour) path to the reduced \uno problem is given. Let the path
have a total length (cost) of $D_{opt}^{\uno}$. Let $s_ig_i$ be the first such
edge that is not in the \uno solution. Because the path is a tour, following
$s_i$ along the path will eventually reach $g_i$. The resulting path will
have the form $s_iv_1\ldots v_2g_iv_3$, i.e., the black path in
Fig.~\ref{fig:uno-hardness}. 
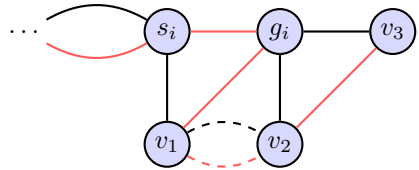
\begin{figure}[htp]
	\begin{center}
	\begin{tikzpicture}[scale=1.5]
		\foreach \nodeName/\nodeLocation in {s_i/{(0, 1)}, v_1/{(0, 0)}, v_2/{(1, 0)}, g_i/{(1, 1)}, v_3/{(2, 1)}}{
			\node (\nodeName) at \nodeLocation {$\nodeName$};
		}
		\node[draw=none, fill=none] (helper) at (-1.25, 1) {$\dots$};
		\foreach \edgeFrom/\edgeTo in {s_i/v_1, g_i/v_2, g_i/v_3}{
			\draw [-] (\edgeFrom) to (\edgeTo);
		}
		\foreach \edgeFrom/\edgeTo in {s_i/g_i, g_i/v_1, v_2/v_3}{
			\draw [-, color=customLightRed] (\edgeFrom) to (\edgeTo);
		}
		\draw [-] (helper) to [out = 30, in = 150] (s_i);
		\draw [-, color=customLightRed] (helper) to [out = -30, in = -150] (s_i);
		\draw [-, dashed] (v_1) to [out = 30, in = 150] (v_2);
		\draw [-, color=customLightRed, dashed] (v_1) to [out = -30, in = -150] (v_2);
	\end{tikzpicture}
	\end{center}
	\caption{\label{fig:uno-hardness} Augmenting a path in an \uno
solution.}  
\end{figure}

Upon the observation of such a partial solution, proceed to make 
the augmentation and replace the path with the new one (red path in 
Fig.~\ref{fig:uno-hardness}). Because $s_ig_i \ll 1/(4n)$, the potential 
increase in path length is bounded by (note that $v_2v_3$ is shorter
than the additive length of $v_2g_i$ and $g_iv_3$)
\[
\|s_ig_i\|_2 + \|g_iv_1\|_2 - \|s_iv_1\|_2 \le 2\varepsilon \ll 1/(2n). 
\]

After at most $n$ such augmentations, an optimal \uno solution is 
converted to an \rno solution. The \rno solution has a cost increase of 
at most $n*1/(2n) = 1/2$. The \rno solution can then be converted
to a solution of the Euclidean-\tsp problem, which will not increase the 
cost. Thus, a \uno solution can be converted to a corresponding 
Euclidean-\tsp solution with a cost addition of less than $1/2$. Let the 
Euclidean-\tsp solution obtained in this manner have a total cost of $D'$, then 
\begin{equation}\label{equ:tuno1}
D' < D^{\uno}_{opt} + \frac{1}{2}.
\end{equation}

Now again let the optimal Euclidean-\tsp solution have a cost of $D_{opt}$.
The solution can be converted to an \rno solution with a total cost
of less than $D_{opt} + 1/4$. The \rno solution is also a solution to 
the \uno problem. That is, for this new \uno solution, the cost is 
\begin{equation}\label{equ:tuno2}
D^{\uno} < D_{opt} + \frac{1}{4}. 
\end{equation}
Now, if $D' > D_{opt}$, then $D' \ge D_{opt} + 1$. Putting this 
together with~\eqref{equ:tuno1} and~\eqref{equ:tuno2}, 
\[
D^{\uno} < D_{opt} + \frac{1}{4} \le D' - \frac{3}{4} \le D^{\uno}_{opt} - \frac{1}{4},
\]
which is a contradiction. Therefore, $D' > D_{opt}$ cannot be true.
Therefore, a cost-optimal \uno solution yields an optimal solution to 
the original Euclidean-\tsp problem. This shows that \uno is at least as 
hard as Euclidean-\tsp.
\end{proof}
\section{Exact ILP-Based Algorithms for Finding Optimal FVS}\label{app:exact}

To compute the exact solution, the problem is modeled as an ILP problem, and 
then solved using LP solvers, e.g., Gurobi \tsp Solver \cite{Gurobi}. In this 
paper, two different ILP models are used, which are similar to the models 
introduced in sections 3.1 and 3.2 of \cite{BahSchNeu15}:
\begin{enumerate}
\item \textbf{ILP-Constraint.} By splitting all vertices $o_i \in G_{dep}$ to $o_i^{in}$ and $o_i^{out}$, a new graph $G_{arc}(V_{arc}, E_{arc})$ is constructed, where $V_{arc} = \{o_1^{in}, o_1^{out}, \dots, o_n^{in}, o_n^{out}\}$, and $(o_i^{out}, o_j^{in}) \in E_{arc}$ \textit{iff} $(o_i, o_j) \in A_{dep}$. By adding extra edges $(o_i^{in}, o_i^{out})$ for all $1 \leq i \leq n$ to $E_{arc}$, problem is transformed to a \textsl{minimum feedback arc set} problem, where the objective is to find a minimum set of arcs to make $G_{arc}$ acyclic. Moreover, every edge in this set ends at $o_i^{in}$ or starts from $o_i^{out}$ can be replaced by $(o_i^{in}, o_i^{out})$, which stands for a vertex $o_i$ in $G_{dep}$, without changing the feasibility of the solution.

The next step is to find a minimum cost ordering $\pi^*$ of the nodes in $G_{arc}$.
Let $c_{i,j} = 1$ if edge $(i,j) \in E_{arc}$, while $c_{i,j} = 0$ if
edge $(i,j) \notin E_{arc}$. Furthermore, let binary variables $y_{i,j}$
associate the ordering of $i,j \in \pi$, where $y_{i,j} = 0$ if $i$
precedes $j$, or $1$ if $j$ precedes $i$. Suppose $|V_{arc}| = m$, the LP formulation is expressed as:
\[\displaystyle \min_{y}~~ \displaystyle \sum_{j = 1}^m (\displaystyle \sum_{k = 1}^{j - 1} c_{k,j} y_{k,j} + \displaystyle \sum_{l = j + 1}^n c_{l,j}(1 - y_{j,l})) \]

\begin{alignat*}{10}
    \textrm{s.t.\quad} & & y_{i,j}   &&+ y_{j,k} &&- y_{i,k} &&\leq 1,\quad  && 1 \leq i < j < k \leq m \\
                       & & - y_{i,j} &&- y_{j,k} &&+ y_{i,k} &&\leq 0,\quad  && 1 \leq i < j < k \leq m \\
\end{alignat*}

The solution arc set contains all the backward edges in $\pi^*$.

\item \textbf{ILP-Enumerate.} First find the set $C$ of all the simple cycles in $G_{dep}$. A set of binary variables $V = \{v_1, \dots, v_n\}$ is defined, each assigned to an object $o_i \in O$, the LP formulation is expressed as::
\begin{equation*}
	\begin{array}{lll}
	\displaystyle \max_{v} & \multicolumn{2}{l}{\displaystyle \sum_{v_i \in V} v_i} \\
	\textrm{s.t.} & \displaystyle \sum_{o_i \in C_j} v_i < |C_j|, &\forall C_j \in C. \\
	\end{array}
\end{equation*}		

Then the vertices in the minimum FVS is the objects whose corresponding variable $v_i$ is 0 in the solution of this LP model.
\end{enumerate}

\section{ILP Model to Find the Shortest Travel Distance in \toro}\label{app:lwolp}

Given a \toro instance with $n$ objects $\objects = \{o_1, \dots, o_n\}$,
without loss of generality, $B = \{o_1, \dots, o_p\} \subset \objects$ 
denotes the set of objects to be moved to buffers, which is calculated by the methods 
introduced in Section~\ref{sec:rwos} and Appendix~\ref{app:exact}. 
The maximum number of buffers to be used is denoted as $p = |B|$.
The ILP model introduced in this section finds the distance-optimal solution amongst all candidates 
that moves objects in $B$ to intermediate locations before moving them to goal configurations while 
employing the minimum number of grasps (i.e. $n + p$).

All variables in this ILP model are boolean variables, 
and have two components: \textsl{nodes} and \textsl{edges}, 
where a node denotes the occupancy of a position on the tabletop at a specific time step, 
and an edge represents a movement of the manipulator between two positions.
The variables appear as a repeated graph pattern in a discrete time domain $t \in \{0, \dots, n + p\}$,
where time step $t$ correlates to the states of the system after the $t$th pick-and-place action.

Assume the following: 
\begin{alignat*}{3}
    1 &\leq i    &&\leq n,\\ 
    1 &\leq j    &&\leq p,\\ 
    1 &\leq k    &&\leq p,\\ 
    1 &\leq \ell &&\leq n,\\ 
    p &< m       &&\leq n.
\end{alignat*}

\subsection{Variables: Nodes} 
There are three kinds of nodes: 
\begin{itemize}
	\item $\{s_1^t, \dots, s_n^t\}$, occupancy of start configurations.
	$s_i^t = 1$ indicates $o_i$ is at its start configuration at time step $t$.
	\item $\{g_1^t, \dots, g_n^t\}$, occupancy of goal configurations.
	$g_i^t = 1$ indicates $o_i$ is at its goal configuration at time step $t$.
	\item $\{b_{11}^t, b_{12}^t, \dots,b_{pp}^t\}$, occupancy of buffers. 
	$b_{jk}^t = 1$ indicates $o_j$ is in buffer $b_k$ at time step $t$.
	\item $s_M, g_M$, indicate the occupancy of the manipulator's rest positions.
\end{itemize}

The value of the nodes when $t = 0$ and $t = n + p$ indicate the start and goal arrangement, respectively. 
Specifically, 
\begin{alignat*}{5}
    s_i^0       & = 1,\quad g_i^0       & & = 0,\quad b_{jk}^0       &&= 0, \\
    s_i^{n + p} & = 0,\quad g_i^{n + p} & & = 1,\quad b_{jk}^{n + p} &&= 0.
\end{alignat*}

\subsection{Variables: Edges}
Edges model the movement of the manipulator and are included in the cost function. 
An edge connects two nodes, e.g. node $1$ and $2$, and is denoted as $e(12)$.
Edges in the ILP model are listed as follows.
\begin{itemize}
	\item $e(s_m^tg_m^t)$, for $1 \leq t \leq n + p$. A positive value indicates a pick-and-place 
	action that brings $o_m$ from its start configuration to its goal configuration at time step $t$.
	\item $e(s_j^tb_{jk}^t)$, $e(b_{jk}^tg_j^t)$, for $1 \leq t \leq n + p$. A positive value 
	indicates a pick-and-place action that brings $o_j$ from its start configuration 
	to buffer $k$, and from buffer $k$ to its goal configuration.
	\item $e(g_i^ts_{\ell}^{t + 1})$, $e(b_{jk}^ts_i^{t + 1})$, $e(g_i^tb_{jk}^{t + 1})$, for $1 \leq t < n + p$, 
	indicates a movement of the manipulator without an object being grasped.
	\item $e(s_Ms_i^1)$, $e(g_i^{n + p}g_M)$, denote the target objects of 
	the first and the last pick-and-place actions.
\end{itemize}

Note that because each edge is associated with a movement of the manipulator, 
the total travel distance of the end-effector can be modeled as 
the summation of the value (i.e. 0 or 1) of all edges multiplied with the cost associated with the edge.
The objective of this ILP model is
\[\min \sum_{e} \text{cost}(e),\]
where $\text{cost}(e)$ denotes the distance between positions 
associated with nodes connected by edge $e$.

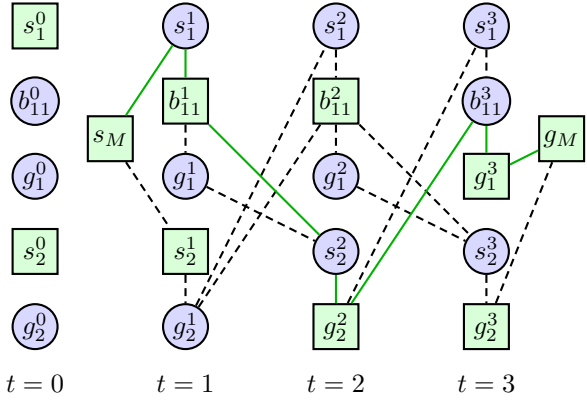
\begin{figure}[htp]
	\centering
	\begin{tikzpicture}[scale=1]
		\foreach \nodeName/\nodeLocation in {
			b_{11}^0/{(-1, 4)}, g_1^0/{(-1, 3)},  g_2^0/{(-1, 1)},
			s_1^1/{(1, 5)}, g_1^1/{(1, 3)}, g_2^1/{(1, 1)},
			s_1^2/{(3, 5)}, g_1^2/{(3, 3)}, s_2^2/{(3, 2)}, 
			s_1^3/{(5, 5)}, b_{11}^3/{(5, 4)}, s_2^3/{(5, 2)}}{
				\node (\nodeName) at \nodeLocation {$\nodeName$};}
		\foreach \nodeName/\nodeLocation in {
			s_M/{(0, 3.5)}, g_M/{(6, 3.5)}, 
			s_1^0/{(-1, 5)}, s_2^0/{(-1, 2)},
			b_{11}^1/{(1, 4)}, s_2^1/{(1, 2)},
			b_{11}^2/{(3, 4)}, g_2^2/{(3, 1)},
			g_1^3/{(5, 3)}, g_2^3/{(5, 1)}}{
				\node[fill=green!15,shape=rectangle] (\nodeName) at \nodeLocation {$\nodeName$};}
		\foreach \nodeName/\nodeLocation in {
			t=0/{(-1, 0.25)}, t=1/{(1, 0.25)}, t=2/{(3, 0.25)}, t=3/{(5, 0.25)}}{
				\node[draw=none, fill=none, minimum size=0, inner sep=0] (\nodeName) at \nodeLocation {$\nodeName$};}
		\foreach \edgeFrom/\edgeTo in {
			s_M/s_2^1, g_2^3/g_M,
			s_2^1/g_2^1, s_2^3/g_2^3,
			s_1^2/b_{11}^2, s_1^3/b_{11}^3,  
			b_{11}^1/g_1^1, b_{11}^2/g_1^2, 
			g_1^1/s_2^2, 
			g_1^2/s_2^3,
			g_2^1/s_1^2, g_2^1/b_{11}^2, 
			g_2^2/s_1^3, 
			b_{11}^2/s_2^3}{
			\draw [-, densely dashed] (\edgeFrom) to (\edgeTo);}
		\foreach \edgeFrom/\edgeTo in {
			s_M/s_1^1, g_1^3/g_M, 
			s_2^2/g_2^2, 
			s_1^1/b_{11}^1, 
			b_{11}^3/g_1^3,
			g_2^2/b_{11}^3,
			b_{11}^1/s_2^2}{
			\draw [-, green!70!black] (\edgeFrom) to (\edgeTo);}
	\end{tikzpicture}
	\caption{
		An example of ILP model when $n = 2, p = 1$ as well as value of variables after solving this model.
		The green square nodes and green solid edges indicate the value of variables is $1$. 
		The solution is interpreted as 
		(i) bring $o_1$ to a buffer, 
		(ii) bring $o_2$ to its goal, 
		(iii) bring $o_1$ to its goal.
		}
	\label{fig:ilp-model}
\end{figure}

An illustration of the ILP model is provided in Fig.~\ref{fig:ilp-model}.

\subsection{Constraints}
This section addresses the constraints that appear in the ILP model.  These
constraints update the value of variables, which ensures that the solution
returned by the ILP model is able to be reduced to a feasible solution for the
original \toro.

Begin by considering the constraints of the first pick-and-place action. 
To update the occupancy of start configurations:
\[\sum_{i = 1}^n e(s_M s_i^1) = s_M = 1, s_i^1 = s_i^0 - e(s_M s_i^1).\]
To update the occupancy of buffers and goal configurations:
\begin{alignat*}{3}
    \sum_{k = 1}^p e(s_j^1 b_{jk}^1) & = e(s_M s_j^1), b_{jk}^1 & & = b_{jk}^0 + e(s_j^1 b_{jk}^1),\\
    e(s_m^1 g_m^1)                   & = e(s_M s_m^1), g_m^1    & & = g_m^0 + e(s_m^1 g_m^1).
\end{alignat*}

After this step, the constraints for all other time steps can be modeled. 
Assume the following:  $1 \leq t \leq n + p - 1$.

The movement of the manipulator between time step $t$ and $t + 1$ correspond to:
\begin{alignat*}{3}
    e(s_j^t b_{jk}^t)                  & = \sum_{i = 1}^n e(b_{jk}^t s_i^{t + 1}) & &+ \sum_{a = 1}^p \sum_{b = 1}^p e(b_{jk}^t b_{ab}^{t+1}), \\
    \sum_{j = 1}^{p} e(b_{jk}^t g_j^t) & = \sum_{i = 1}^n e(g_j^t s_i^{t + 1})    & &+ \sum_{k = 1}^p \sum_{l = 1}^p e(g_j^t b_{kl}^{t+1}), \\
    e(s_m^t g_m^t)                     & = \sum_{i = 1}^n e(g_m^t s_i^{t + 1})    & &+ \sum_{k = 1}^p \sum_{l = 1}^p e(g_j^t b_{kl}^{t+1}).
\end{alignat*}

For time step $t+1$, constraints are imposed on the incoming edges from time step $t$, 
in order to avoid the scenario where the manipulator travels to a vacant location:
\begin{alignat*}{3}
    \sum_{i = 1}^n e(g_i^t s_{\ell}^{t+1}) & + \sum_{j = 1}^p \sum_{k = 1}^p e(b_{jk}^t s_{\ell}^{t+1}) & & \leq s_{\ell}^t,\\
    \sum_{i = 1}^n e(g_i^t b_{jk}^{t+1})   & + \sum_{a = 1}^p \sum_{b = 1}^p e(b_{ab}^t b_{jk}^{t+1})   & & \leq b_{jk}^t.
\end{alignat*}

Update the edges to simulate a pick-and-place action in time step $t + 1$:
\begin{alignat*}{3}
    \sum_{k = 1}^p e(s_j^{t+1} b_{jk}^{t+1}) & = \sum_{i = 1}^n e(g_i^t s_j^{t+1})    & & + \sum_{a = 1}^p \sum_{b = 1}^p e(b_{ab}^t s_j^{t+1}),\\
    e(b_{jk}^{t+1} g_j^{t+1})                & = \sum_{i = 1}^n e(g_i^t b_{jk}^{t+1}) & & + \sum_{a = 1}^p \sum_{b = 1}^p e(b_{ab}^t b_{jk}^{t+1}),\\
    e(s_m^{t+1} g_m^{t+1})                   & = \sum_{i = 1}^n e(g_i^t s_m^{t+1})    & & + \sum_{j = 1}^p \sum_{k = 1}^p e(b_{jk}^t s_m^{t+1}).
\end{alignat*}

Update nodes in time step $t + 1$:
\begin{alignat*}{3}
    s_i^{t + 1}  & = s_i^t & &- \sum_{\ell = 1}^n e(g_{\ell}^t s_i^{t+1}) - \sum_{j = 1}^p \sum_{k = 1}^p e(b_{jk}^t s_i^{t+1}), \\
    b_{jk}^{t+1} & = b_{jk}^t & & + e(s_j^{t+1} b_{jk}^{t+1}) \\
                 & & & - \sum_{i = 1}^n e(g_i^t b_{jk}^{t+1}) - \sum_{a = 1}^p \sum_{b = 1}^p e(b_{ab}^t b_{jk}^{t+1}),
\end{alignat*}

\begin{alignat*}{3}
    g_j^{t + 1} &= g_j^t & &+ \sum_{k = 1}^p  e(b_{jk}^{t+1} g_j^{t+1}),\\
    g_m^{t + 1} &= g_m^t & &+ e(s_m^{t + 1} g_m^{t+1}).
\end{alignat*}

Since each buffer is presented as multiple copies in each time step, 
one must make sure it is occupied by at most one object:
\[\sum_{j = 1}^p b_{jk}^t \leq 1.\]

Update the dependencies in the dependency graph.
Suppose $s_i$ is in collision with $g_{\ell}$ then
\[s_i^t + g_{\ell}^t \leq 1.\]

After employed all $n + p$ pick-and-place actions, 
the manipulator goes to the rest position $g_M$:
\begin{alignat*}{3}
    e(g_j^{n + p} g_M) &= \sum_{k = 1}^p e(b_{jk}^{n + p} g_j^{n + p}), \\
    e(g_m^{n + p} g_M) &= e(s_m^{n + p} g_m^{n + p}), \\
    g_M                &= \sum_{j = 1}^p e(g_j^{n + p} g_M) + \sum_{m = p + 1}^{n} e(g_m^{n + p} g_M) = 1.
\end{alignat*}

\section{Solution for an \rno Instance in Physical Experiment Section}\label{app:exp-sol}

\begin{figure}[!ht]
    \begin{tabularx}{\linewidth}{>{\centering\arraybackslash}X>{\centering\arraybackslash}X} 
        \includegraphics[width = 0.99 \linewidth]{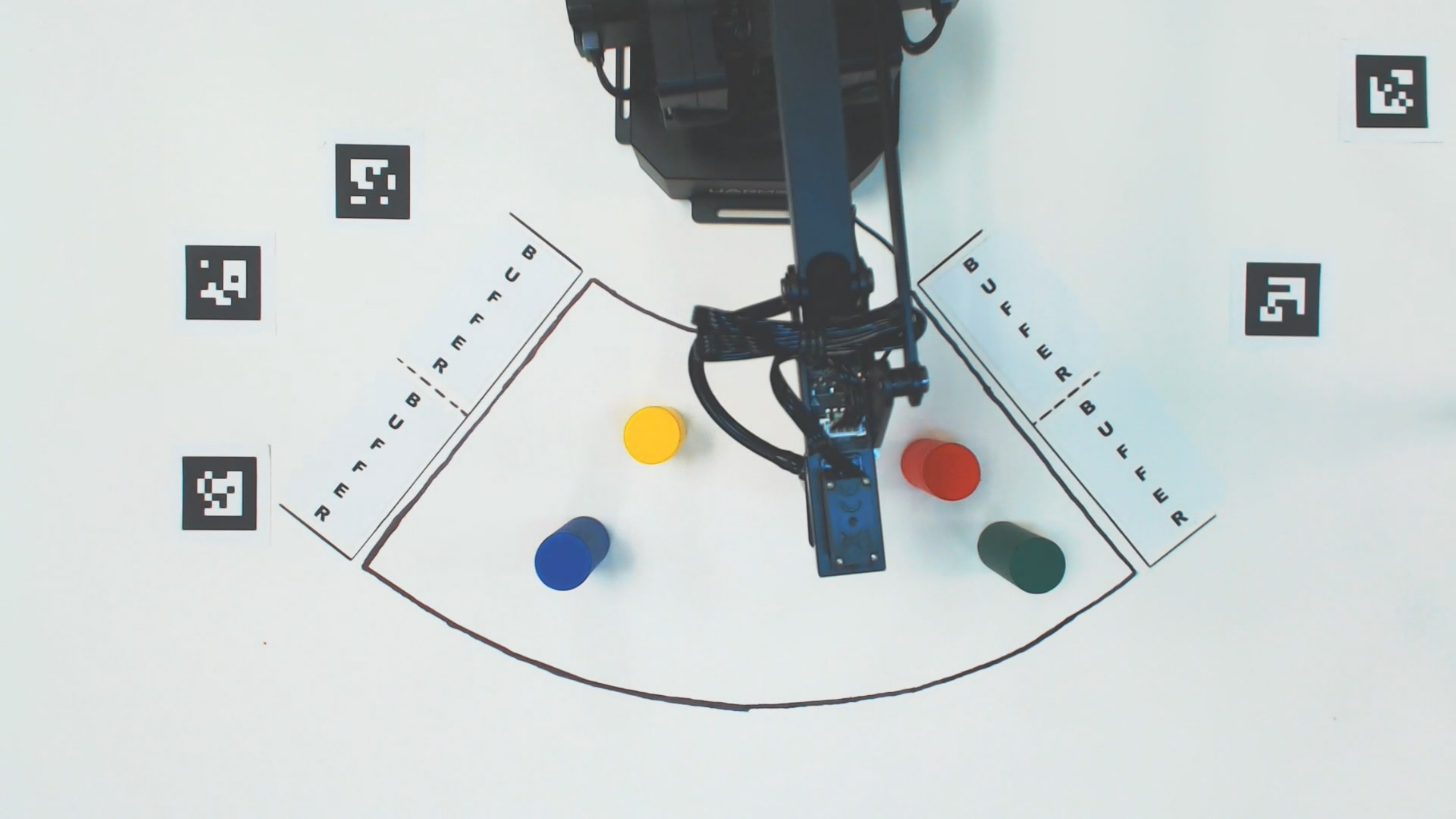} &
        \includegraphics[width = 0.99 \linewidth]{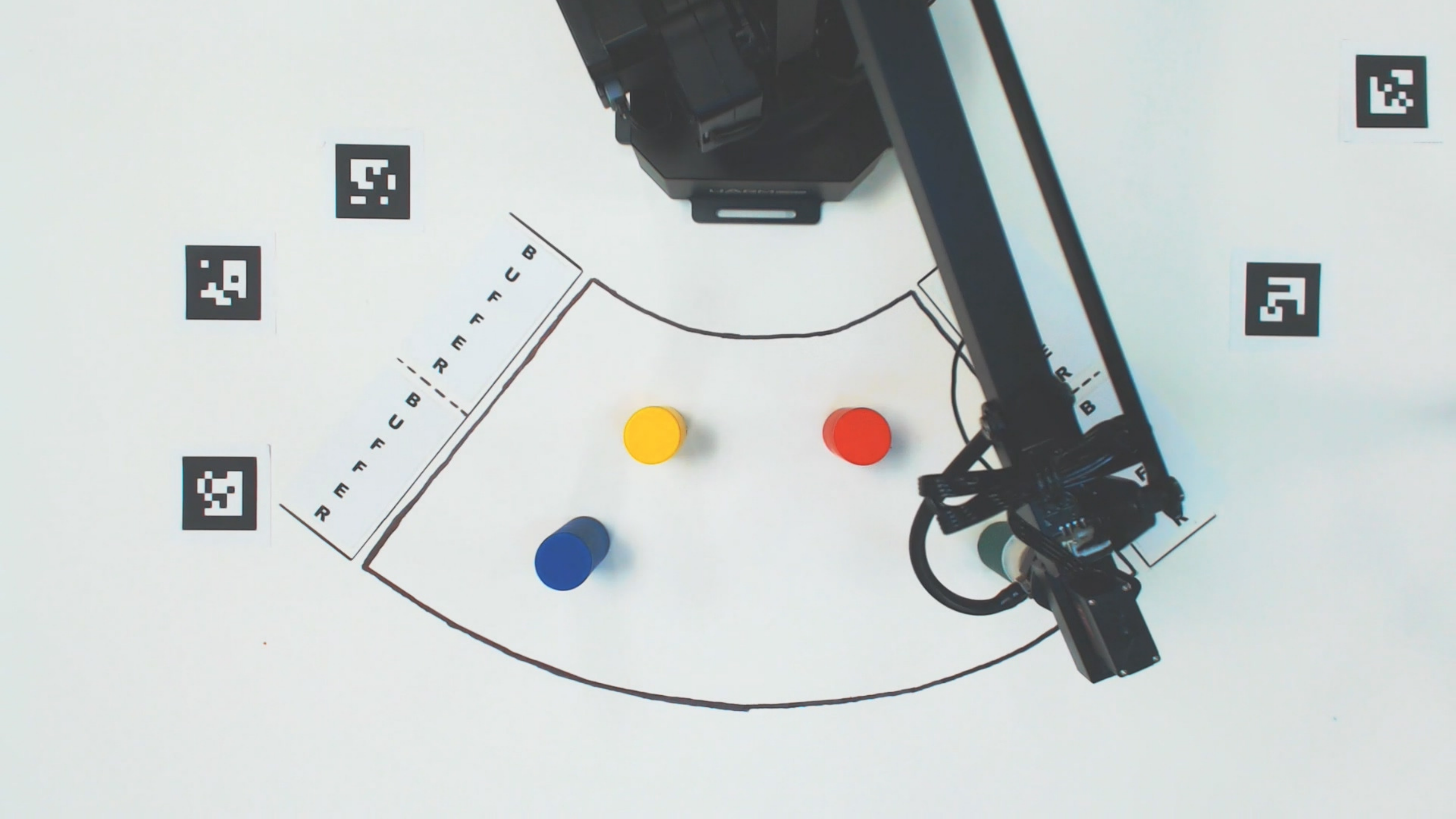} \\
        (a) & (b) \\
        \includegraphics[width = 0.99 \linewidth]{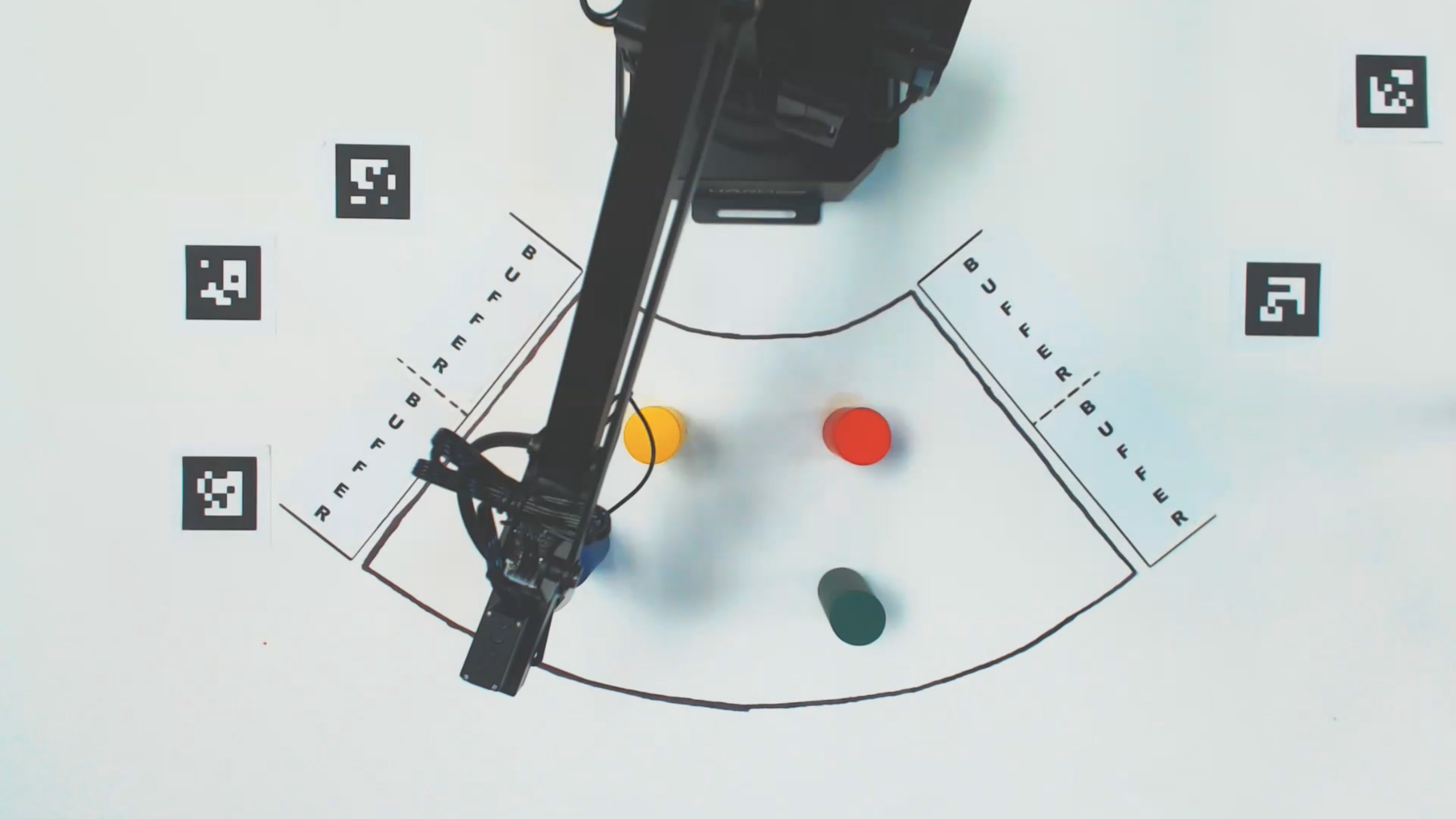} &
        \includegraphics[width = 0.99 \linewidth]{scenario-1-goal.pdf} \\
        (c) & (d) \\
    \end{tabularx}
    \caption{
        Step by step solution for the \rno instance in Fig.~\ref{fig:experiment-1}.}
    \label{fig:exp-solution}
\end{figure}

Fig.~\ref{fig:exp-solution} demonstrates the optimal solution for 
the problem instance shown in Fig.~\ref{fig:experiment-1}. 
The sequence of pick-and-place actions over the colored cylinders 
are yellow, red, green, blue. 

\section*{Acknowledgment}
    We thank uFactory for supplying the uArm Swift Pro robot that was used in the hardware-based evaluation.\\
    
    \noindent The authors would like to recognize the contributions of Jiakun Lyu \{Zhejiang University, China\} for his preliminary work on the object pose detection used in the hardware experiments.

    This work is supported by NSF awards IIS-1617744,
    IIS-1451737 and CCF-1330789, as well as internal support by Rutgers University.
    Any opinions or findings expressed in this paper do not necessarily reflect the
    views of the sponsors. 

    A video containing complete executions for each of the 
    three scenarios described in the Experimental Validation 
    (Section~\ref{ssect:ev}) can be found at {\tt https://youtu.be/Ub7QSDQz0Qk}.

\bibliographystyle{IEEETran}
\bibliography{master}
  

\end{document}